\documentclass[10pt,twocolumn,letterpaper]{article}

\usepackage[pagenumbers]{wacv} 

\usepackage{graphicx}
\usepackage{amsmath}
\usepackage{amssymb}
\usepackage{booktabs}
\usepackage{tabularx}
\usepackage{multirow}

\graphicspath{{./figures}}
\DeclareGraphicsExtensions{.pdf,.jpeg,.jpg,.png}

\newcolumntype{Y}{>{\centering\arraybackslash}X}
\newcolumntype{s}{>{\hsize=.5\hsize}Y}
\newcolumntype{d}{>{\hsize=.5\hsize}X}
\newcolumntype{e}{>{\hsize=.5\hsize}l}

\usepackage[pagebackref,breaklinks,colorlinks]{hyperref}

\usepackage[capitalize]{cleveref}
\crefname{section}{Sec.}{Secs.}
\Crefname{section}{Section}{Sections}
\Crefname{table}{Table}{Tables}
\crefname{table}{Tab.}{Tabs.}

\def\wacvPaperID{1211} 
\def\confName{WACV}
\def\confYear{2025}

\begin{document}

\title{Global-Local Similarity for Efficient Fine-Grained Image Recognition
\\with Vision Transformers}

\author{
Edwin Arkel Rios\textsuperscript{\dag}, Min-Chun Hu\textsuperscript{\ddag}, Bo-Cheng Lai\textsuperscript{\dag}\\
\vspace{-0.2cm}
\textsuperscript{\dag}\textit{National Yang Ming Chiao Tung University, Taiwan}, \textsuperscript{\ddag}\textit{National Tsing Hua University, Taiwan}\\
}
\maketitle

\begin{abstract}
    Fine-grained recognition involves the classification of images from subordinate macro-categories, and it is challenging due to small inter-class differences. To overcome this, most methods perform discriminative feature selection enabled by a feature extraction backbone followed by a high-level feature refinement step. 
    Recently, many studies have shown the potential behind vision transformers as a backbone for fine-grained recognition, but their usage of its attention mechanism to select discriminative tokens can be computationally expensive.
    In this work, we propose a novel and computationally inexpensive metric to identify discriminative regions in an image. We compare the similarity between the global representation of an image given by the CLS token, a learnable token used by transformers for classification, and the local representation of individual patches. We select the regions with the highest similarity to obtain crops, which are forwarded through the same transformer encoder. Finally, high-level features of the original and cropped representations are further refined together in order to make more robust predictions.
    Through extensive experimental evaluation we demonstrate the effectiveness of our proposed method, obtaining favorable results in terms of accuracy across a variety of datasets. Furthermore, our method achieves these results at a much lower computational cost compared to the alternatives. Code and checkpoints are available at: \url{https://github.com/arkel23/GLSim}.

\end{abstract}

\section{Introduction}
\label{sec_introduction}

Fine-grained image recognition (FGIR) involves classifying sub-categories within a larger super-category. Examples of FGIR problems include differentiating between bird species \cite{van_horn_building_2015, wah_caltech-ucsd_nodate} and anime characters \cite{noauthor_tagged_nodate, wang_danbooru_2019}. It is a widely studied area with various applications such as automatic biodiversity monitoring, intelligent retail, among others \cite{wei_fine-grained_2021}. However, FGIR is a challenging task due to small inter-class differences and large intra-class variations.

In order to tackle these challenges, most of the existing methods equip a coarse image recognition backbone encoder with modules to select discriminative regions \cite{liu_filtration_2020, yang_learning_2018}. Recently, with the advent of vision transformers (ViTs) \cite{dosovitskiy_image_2020}, many researchers have explored using this new architecture as a backbone for fine-grained image recognition \cite{he_transfg_2022, wang_feature_2021, zhang_free_2022}. Since transformer encoder blocks utilize multi-head self-attention, a natural choice was to use the attention scores to guide the discriminative region selection process, eliminating the need for an external attention module \cite{hu_see_2019, rao_counterfactual_2021, zhuang_learning_2020}. These regions are either cropped and re-inputted into the same encoder \cite{hu_rams-trans_2021, zhang_free_2022}, or combined using high-level feature refinement modules \cite{he_transfg_2022, wang_feature_2021}, or both \cite {ji_dual_2023}, before making final predictions.

However, as seen in \Cref{table_model_comparison_b16}, several methods leveraging matrix-multiplication for attention aggregation \cite{abnar_quantifying_2020, conde_exploring_2021, hu_rams-trans_2021, zhu_dual_2022, he_transfg_2022, liu_transformer_2022} exhibit significant computational expense, characterized by a complexity of $\mathcal{O}(N^3)$ with respect to the sequence length $N$. Notably, the computational cost associated with attention aggregation can exceed that of the backbone's forward pass, particularly as image size increases. This computational burden presents a substantial limitation for FGIR tasks, which often benefit from the use of higher resolution images \cite{guo_fine-grained_2023, wang_efficient_2023, sun_fair1m_2022}, thereby constraining the practical applicability of these methods.

To address this, we introduce a novel metric, \textbf{GLS (Global-Local Similarity)}, to identify discriminative regions for Vision Transformers (ViTs) with \textbf{a computational cost several orders of magnitude lower than that of aggregated attention}. We compute the similarity between the global representation of an image, as given by the ViT's CLS token typically used for classification, and the local representations of individual patches. Regions exhibiting high similarity in the high-dimensional feature space are presumed to share underlying factors influencing the global representation, rendering them highly representative of the image. We subsequently crop the image based on the regions with highest GLS, resize it, and re-input it into the encoder. Finally, the high-level features of the original and cropped image image are refined collectively using an aggregator module, enhancing the robustness of predictions. Our contributions are summarized as follows:

\begin{itemize}
    \item We propose a novel metric for ViTs to identify discriminative regions in an image. GLS can be used as a visualization tool for interpretability of classification decisions, does not require any additional parameters and exhibits linear complexity $\mathcal{O}(N)$ with respect to sequence length. Compared to commonly used matrix-multiplication based aggregated attention the computational cost of GLS can be from $\mathbf{10^3}$ up to $\mathbf{10^6}$ times lower, depending on the architecture and image size.
    \item We incorporate the proposed metric into a method that selects discriminative crops from an image and aggregates high-level features from both the original and cropped image for improved fine-grained image recognition performance.
    \item We conducted a thorough analysis of fine-grained recognition models by comparing models across 10 datasets spanning a wide spectrum of tasks. Our model achieves the highest accuracy in 8 datasets, and on average, reduces the relative classification error\footnote{Relative error: $Err_{rel} = 100 \cdot \dfrac{(100 - Acc) - (100 - Acc_{ref})}{100 - Acc_{ref}}$} by \textbf{10.15\%} compared to the baseline ViT. These results demonstrate the potential of the proposed global-local similarity as discriminative region selection criteria. Moreover, our model achieves these results with \textbf{9.26x} less VRAM and a \textbf{2.59x} higher inference throughput than the best performing model in the other 2 datasets.
\end{itemize}


\section{Related Work}
\label{sec_related_work}


\subsection{Fine-Grained Image Recognition}
\label{ssec_fine_grained}

To address challenges of intra-class variations in FGIR, most approaches aim to identify discriminative regions that encapsulate the subtle differences between classes. Initially, part-level bounding boxes \cite{zhang_part-based_2014} or segmentation masks \cite{wei_mask-cnn_2018} were employed to train localization subnetworks, but the cost of manual annotations limited their applicability to a wide variety of tasks.

Therefore, researchers aimed to use weak supervision, i.e., image-level labels, to localize discriminative regions. In this category most methods utilize either RPN \cite{ren_faster_2015} or attention mechanism for this goal. NTS-Net \cite{yang_learning_2018} is an example of the former that leverages classifier confidence to encourage collaboration between a Navigator and a Teacher network to propose and evaluate informative regions. RA-CNN \cite{fu_look_2017},  WS-DAN \cite{hu_see_2019} and CAL \cite{rao_counterfactual_2021} are examples of the latter approach. RA-CNN employs an attention proposal subnetwork that recursively proposes finer regions, while WS-DAN and CAL use attention maps to generate augmentations of the image. While these methods achieve competitive results, they rely on external (attention) modules that increase the computational cost. Moreover, they require multiple recursive stages or multiple crops to obtain high-quality predictions, which further exacerbates the computational cost and limits their practicality.

\begin{table*}[!htb]
    \centering

    \centering

    \small{
    \begin{tabularx}{\linewidth}{X X X X X X}
        \toprule
        Model & TransFG \cite{he_transfg_2022} & Various \cite{hu_rams-trans_2021, zhu_dual_2022, liu_transformer_2022} & FFVT \cite{wang_feature_2021} & AF-Trans \cite{hu_rams-trans_2021} & Ours \\
        \midrule
        DFSM & PSM & Rollout & MAWS & SACM & GLS \\
        Complexity & $L\cdot H \cdot N^3$ & $L\cdot N^3$ & $L\cdot H\cdot N$ & $L\cdot H \cdot N^2$ & $N\cdot D$ \\
        \midrule

        FLOPs (IS=224) & $8.2\times 10^3$ & $1.7\times 10^2$ & $1.3\times 10^-1$ & $1.4\times 10^1$ & $4.5\times 10^-1$ \\
        \% of Backbone & 46.531 & 0.9833 & 0.0007 & 0.0795 & 0.0026 \\
        FLOPs (IS=448) & $6.2\times 10^5$ & $1.1\times 10^4$ & $5.3\times 10^-1$ & $2.5\times 10^2$ & $1.8\times 10^0$ \\
        \% of Backbone & \textbf{781.31} & 13.592 & 0.0007 & 0.3162 & 0.0023 \\
        FLOPs (IS=768) & $1.5\times 10^7$ & $2.7\times 10^5$ & $1.5\times 10^0$ & $2.1\times 10^3$ & $5.3\times 10^0$ \\
        \% of Backbone & \textbf{5065.4} & \textbf{91.152} & 0.0005 & 0.7072 & 0.0018 \\
        FLOPs (IS=1024) & $9.1\times 10^7$ & $1.5\times 10^6$ & $2.7\times 10^0$ & $6.9\times 10^3$ & $9.4\times 10^0$ \\
        \% of Backbone & \textbf{13674} & \textbf{228.84} & 0.0004 & 1.0487 & 0.0014 \\

        \bottomrule
    \end{tabularx}
    }

    \caption{Summary of computational cost for discriminative feature selection mechanisms (DFSM) used by ViTs in FGIR and their ratio of FLOPs compared to the backbone for ViT B-16. We highlight in \textbf{bold} values that exceed 50\% of the FLOPs of the backbone.}
    \label{table_model_comparison_b16}
\end{table*}

\subsection{Transformers for Fine-Grained Recognition}
\label{ssec_vit_fgir}

Recently, with the emergence of transformers in vision, there has been a substantial amount of research in how to effectively exploit this new architecture for FGIR \cite{he_transfg_2022, wang_feature_2021, hu_rams-trans_2021}. The global receptive field of the self-attention mechanism, coupled with the inherent relative importance imparted by the attention weights, makes the ViT a natural candidate for usage as a backbone for fine-grained tasks.

In \Cref{table_model_comparison_b16} we compare the cost of the discriminative feature selection modules (DFSM) of proposed ViTs for FGIR as we observe it is the largest difference between these methods. All evaluated methods make use of the ViT's attention mechanism, mostly by leveraging recursive layer-wise matrix-matrix multiplication (TransFG \cite{he_transfg_2022}, RAMS-Trans \cite{hu_rams-trans_2021}, DCAL \cite{zhu_dual_2022}, TPSKG \cite {liu_transformer_2022}, EV \cite{conde_exploring_2021}) to calculate aggregated attention scores. RAMS-Trans and DCAL first compute the mean over different heads, while TransFG computes these scores separately for each head. AFTrans \cite{zhang_free_2022} computes head-wise aggregation of attention via element-wise multiplication, and re-weights layer contributions via its Squeeze-and-Excitation-like mechanism. FFVT \cite{wang_feature_2021} selects features on a layer-by-layer basis based on the normalized vector product of the first row and column of the attention matrix.

Nevertheless, methods involving matrix-matrix multiplications incur a high computational cost that can even surpass the cost of the backbone itself, specially as the image size increases. The number of FLOPs required for multiplying two matrices $\mathbf{M_1}\in \mathbb{R}^{N\times N}$ and $\mathbf{M_2}\in \mathbb{R}^{N\times N}$ is $N\cdot N \cdot (2N - 1)$. For TransFG, which entails $H$ heads, and requires $L-2$ multiplications in the process of the PSM, its computational complexity is $\mathcal{O}(L\cdot H \cdot N^3)$. Those employing attention rollout \cite{abnar_quantifying_2020} such as RAMS-Trans average the heads, thereby reducing the complexity by a factor of $H$. Although AF-Trans does not perform matrix-matrix multiplication, it still recursively computes $N^2$ element-wise products for $H$ heads in each layer. The computationally lightest DFSM is FFVT's MAWS since it does not involve large matrix products. However, it exhibits considerable variation in performance across tasks, as observed from the results in \Cref{sec_results} and poor interpretability as observed from the visualization in \Cref{figure_vis_dfsm}.

\subsection{Similarity in Computer Vision}
\label{ssec_similarity}

Determining similarity between visual data is a fundamental requirement in various computer vision applications, including image retrieval and matching \cite{shechtman_matching_2007, diao_similarity_2021}. Typically, deep feature extraction models are employed to learn a high-level feature space, where metrics such as cosine similarity, L1, or L2 distance facilitate target retrieval given a query. Examples include CLIP and CCFR. CLIP \cite{radford_learning_2021} achieves this by training image and text encoders to optimize cosine similarity for correct image-text pairs and minimize it for incorrect pairs. CCFR \cite{yang_re-rank_2021} leverages cosine similarity to retrieve external features from a database to re-rank fine-grained classification predictions. Our approach, however, is distinct in that it is more closely related to self-similarity \cite{shechtman_matching_2007}, computing similarity within a single image rather than against external images.

Self-similarity is commonly used as a descriptor that reveals the structural layout of an image or video by measuring similarities of a local patch within its neighborhood \cite{shechtman_matching_2007, junejo_cross-view_2008}. Recently, self-similarity has been harnessed as an intermediate feature transformation in deep neural networks, demonstrating its efficacy in video understanding \cite{kwon_learning_2021}, image translation \cite{zheng_spatially-correlative_2021}, visual correspondence \cite{kim_fcss_2019}, few-shot image recognition \cite{kang_relational_2021}, and image retrieval \cite{lee_revisiting_2023}. 

While self-similarity typically involves transformations based on local patch neighborhoods, our approach computes similarity between the global image representation and individual patches. As a result, our method deviates from conventional self-similarity, which focuses on local patch relations, and instead employs the similarity between the global representation and local features as a metric for discriminative feature selection.


\section{Proposed Method: GLSim}
\label{sec_method_glsim}


\begin{figure*}[!htb]
\centering
\includegraphics[width=1.0\linewidth]{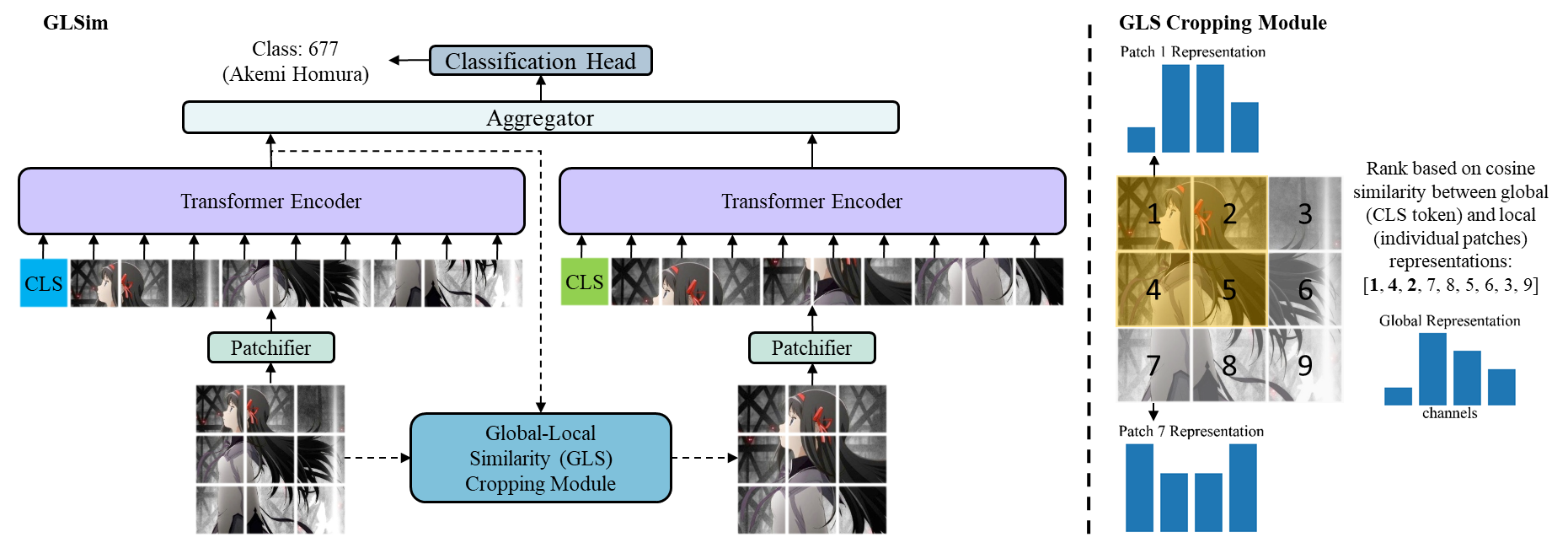}
\caption{Overall flow of our proposed system, GLSim. Starting from the bottom left corner, an image is patchified and passed through a series of transformer encoder blocks to extract features. These are then used by the GLS Module to select discriminative crops, as indicated by the dashed-lines. The GLS Module crops the image according to the coordinates corresponding to the top-$O$ tokens with the highest similarity between global and local representations of the encoded image. The cropped image is then passed through the same encoder. Finally, high-level features from the original and cropped image are refined using an Aggregator module before making the final predictions.}
\label{figure_glsim_overview}
\end{figure*}


An overview of our method is shown in \Cref{figure_glsim_overview}. Images are encoded using a transformer encoder. Then, to find discriminative regions, we employ the Global-Local Similarity (GLS) Module. The GLS Module computes the similarity between the global representation of the image and local tokens and crops the image based on the tokens with highest similarity. The image is then resized and forwarded through the encoder. Finally, an Aggregator module is employed to collectively refine high-level features from the original and cropped image, before being forwarded through a classification head to make our final predictions.

\subsection{Image Encoding with Vision Transformer}
\label{ssec_encoder_vit}

We encode images using a ViT \cite{dosovitskiy_image_2020} encoder. Images are patchified using a convolution with large kernel size $P$ and flattened into a 1D sequence of $D$ channels and length $N=(S_1/P)\times(S2/P)$, where $S_1$ and $S_2$ represent the image width and height. Inspired by BERT \cite{devlin_bert_2019}, the authors concatenate a learnable CLS token at the start of the sequence. Learnable positional embeddings are added to the sequence to incorporate spatial information. This sequence is forwarded through a series of $L$ transformer encoder blocks \cite{vaswani_attention_2017} which apply multi-head self-attention (MHSA) with $H$ heads and position-wise feed-forward networks (PWFFN), before being forwarded through a Layer Normalization \cite{ba_layer_2016} layer.  This output of the transformer is denoted as $\mathbf{f} \in \mathbb{R}^{(N+1) \times D}$.

\subsection{Discriminative Feature Selection With GLS}
\label{sec_gls}

To identify the discriminative regions in the image, we then compute the similarity between the global representation of the image, as given by the CLS token ($\mathbf{f}^0$) and each of the other tokens in the sequence. This similarity map, denoted as $\mathbf{s} \in \mathbb{R} ^ {N\times D}$, is calculated according to \Cref{eq_sim,eq_cos_sim} when cosine similarity is employed as similarity measure:

\begin{equation}
\label{eq_sim}
\mathbf{s}^i = sim(\mathbf{f}^0, \mathbf{f}^i) = cos(\mathbf{f}^0, \mathbf{f}^i) \quad i \in 1, 2, ..., N
\end{equation}

\begin{equation}
\label{eq_cos_sim}
cos(\mathbf{f}^0, \mathbf{f}^i) = \dfrac{\mathbf{f}^0 \cdot \mathbf{f}^i}{\Vert \mathbf{f}^0 \Vert \Vert \mathbf{f}^i \Vert} = \dfrac{\sum_{j=1}^D f^0_j f^i_j}{\sqrt{\sum_{j=1}^D f^0_j} \sqrt{\sum_{j=1}^D f^i_j}}
\end{equation}

Then, we crop the image based on the image coordinates corresponding to a rectangle that encloses the top-$O$ tokens with highest similarity. Subsequently, the cropped image is resized to the original image size $S_1 \times S_2$ before being forwarded through the encoder to obtain $\mathbf{f}_{crops}$. While the encoder's weights are shared, a different CLS token is utilized for the crops, drawing inspiration from \cite{hu_rams-trans_2021}.


\begin{figure*}[!htb]
    \centering
    \includegraphics[width=0.9\linewidth]{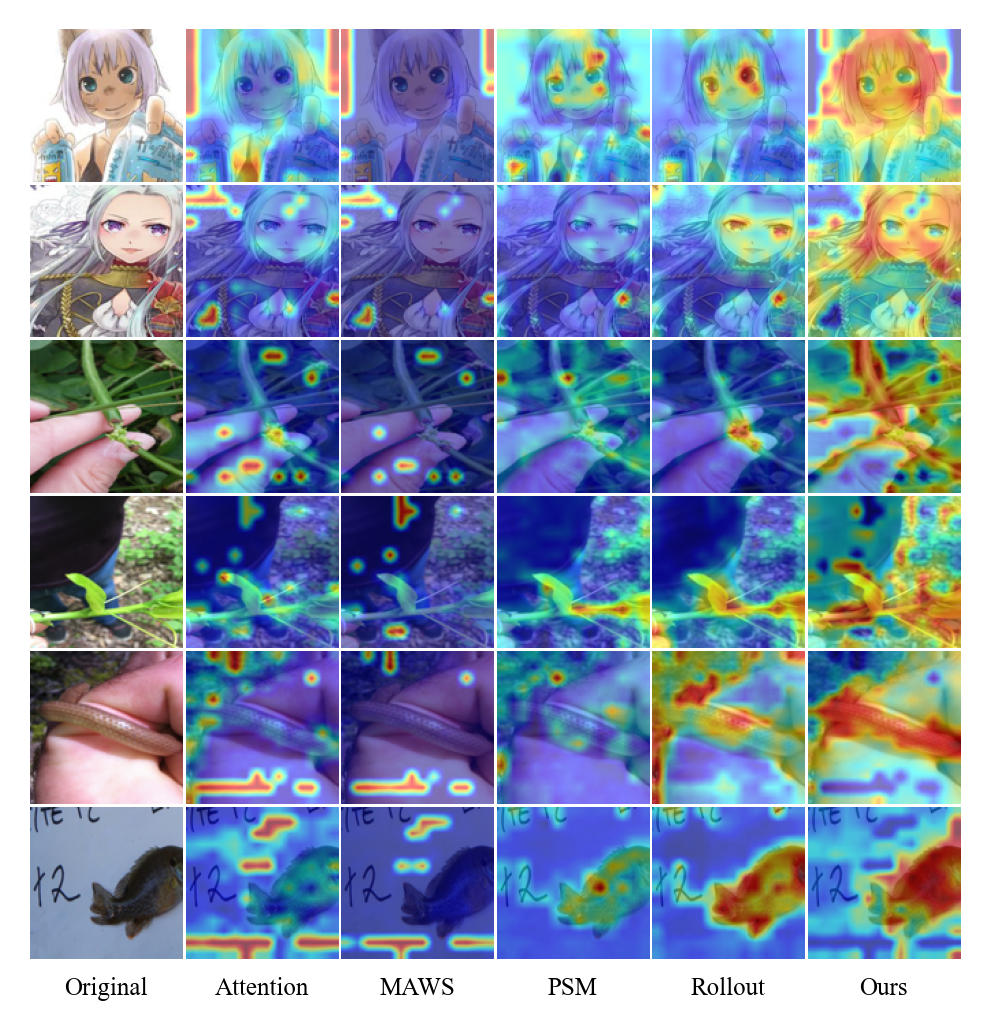}
    \caption{Visualization of discriminative feature selection mechanisms for a fine-tuned ViT B-16 on DAFB \cite{branwen_danbooru2021_2015, rios_anime_2022} (first 2 rows) and iNat17 \cite{van_horn_inaturalist_2018} (last 4 rows) dataset. From left to right: original image, head-wise average of attention scores of last layer, MAWS \cite{wang_feature_2021} of last layer, PSM \cite{he_transfg_2022}, attention rollout \cite{abnar_quantifying_2020}, and our proposed global-local similarity.}
    \label{figure_vis_dfsm}

\end{figure*}


As the CLS token aggregates the discriminative details from the image through self-attention, we expect the local tokens with high degree of similarity in the high-dimensional feature space to share the same underlying factors that drive the global representation. To verify our assumptions, we visualize various discriminative feature selection mechanisms (DFSMs) in \Cref{figure_vis_dfsm}. We note that single-layer attention \cite{wang_feature_2021}, as depicted in the second and third column of the figure, does not focus on the objects of interest in certain scenarios. Conversely, while aggregating multiple layers attention through recursive matrix multiplication (fourth and fifth column) may offer a more effective alternative, it comes at a significant computational cost, up to 13,674x and 229x higher than the forward pass of the backbone, as seen in \Cref{table_model_comparison_b16}. On the other side, while heatmaps of global-local similarity (last column) may not exhibit the same level of focus on specific regions as the aggregated attention maps, they are still mostly aligned with the regions of interest and the computational cost is much lower as seen in \Cref{table_model_comparison_b16}. In particular, GLS requires between $\mathbf{10^3}$ to $\mathbf{10^7}$ times \textbf{less FLOPs} compared to attention rollout \cite{abnar_quantifying_2020} and TransFG's PSM \cite{he_transfg_2022}, making it a highly efficient and effective alternative to attention.

\subsection{High-Level Feature Refinement}

In order to achieve more robust predictions, we explicitly combine high-level features from both the original and cropped images by incorporating an Aggregator module comprised of a single transformer encoder block. Specifically, we concatenate the output CLS token of the original image ($\mathbf{f}^0$) and the cropped image ($\mathbf{f}_{crops}^0$), and forward this concatenated representation through the Aggregator module. This is shown in \Cref{eq_glsim_mhsa,eq_glsim_pwffn}:

\begin{equation}
\label{eq_glsim_mhsa}
\mathbf{r}^\prime = \text{MHSA}(\text{LN}([\mathbf{f}^0; \mathbf{f}_{crops}^0]) + [\mathbf{f}^0; \mathbf{f}_{crops}^0]).
\end{equation}
\begin{equation}
\label{eq_glsim_pwffn}
\mathbf{r} = \text{PWFFN}(\text{LN}(\mathbf{r}^\prime) + \mathbf{r}^\prime
\end{equation}
%


Then we pass these tokens through a LayerNorm layer. We forward the first token in the sequence through a linear layer $C_{final} \in \mathbb{R}^{D \times T}$ to obtain our final classification predictions. We utilize cross-entropy as our loss function. This process is described by \Cref{eq_glsim_ln,eq_glsim_out}:

\begin{equation}
\label{eq_glsim_ln}
\mathbf{r}_{LN} = \text{LN}(\mathbf{r})
\end{equation}
\begin{equation}
\label{eq_glsim_out}
y_{final} = \mathbf{r}_{LN}^0\mathbf{C}_{final}
\end{equation}


\section{Datasets and Experimental Setup}
\label{ssec_experiments}

We conduct experiments on 10 datasets, whose detailed statistics are shown in the Appendix. We perform a learning rate (LR) search for each dataset-model pair with a small split from the train set. Then as suggested by Gwilliam et al. \cite{gwilliam_fair_2021} we report mean accuracy and its standard deviation on the test set across different seeded runs. The best and second best results, when applicable, are highlighted by \textbf{boldface} and \underline{underline}, respectively.


We employ the ViT B-16 \cite{dosovitskiy_image_2020} pre-trained on ImageNet-21k as the backbone for our proposed model. The hyperparameter $O$, which determines how many tokens with highest similarity we use for cropping, is set to 8 for most datasets, except for Dogs where it is set to 16. This value is doubled accordingly when using image size 448. We compare our proposal against several existing state-of-the-art (SotA) models, including the baseline ViT, two other ViTs specifically designed for FGIR (namely, TransFG \cite{he_transfg_2022} and FFVT \cite{wang_feature_2021}), as well as two CNN-based approaches, namely ResNet-101 \cite{he_deep_2015}, and CAL \cite{rao_counterfactual_2021} which employs the former as a backbone. The source code provided by the authors was incorporated into our training and evaluation codebase with minor modifications.

We resize images to a square of size $1.34S_1\times 1.34S_2$ (e.g., 300x300 for image size 224x224) then perform a random crop during training or a center crop at inference. During training we additionally apply random horizontal flipping, random erasing \cite{zhong_random_2020}, trivial augment \cite{muller_trivialaugment_2021}, label smoothing \cite{szegedy_rethinking_2016} and stochastic depth \cite{huang_deep_2016}.

We employ the SGD optimizer with batch size of 8, weight decay of 0 and cosine annealing \cite{loshchilov_sgdr_2017} with a warmup of 500 steps for learning rate scheduling. Models are trained for 50 epochs. We use PyTorch \cite{paszke_pytorch_2019} with mixed precision and employ Weight and Biases \cite{biewald_experiment_2020} for logging and experiment tracking. We conduct all experiments using a single V100 GPU.


\section{Results and Discussion}
\label{sec_results}

\subsection{FGIR SotA Model Comparison}

\subsubsection{Comparison on NABirds}

In the NABirds \cite{van_horn_building_2015} dataset the accuracies of most ViT-based methods are higher compared to CNN-based methods. This could be due to the usage of the self-attention operator with its global receptive field which can allow for effective aggregation of discriminative features from diverse regions of the image. Our method outperforms the best CNN-based method by 1\%. Compared to ViT-based methods, our method outperforms the baseline ViT by 3.1\%, while the second best, Dual-TR \cite{ji_dual_2023} by 0.7\%.

\begin{table}[htb]
    \centering
    \begin{tabularx}{\linewidth}{X X Y}
        \toprule
        Method & Backbone & Top-1 (\%) \\
        \midrule
        MaxEnt \cite{dubey_maximum-entropy_2018} & DenseNet-161 & 83.0 \\
        Cross-X \cite{luo_cross-x_2019} & ResNet-50 & 86.4 \\
        API-Net \cite{zhuang_learning_2020} & ResNet-101 & 88.1 \\
        CS-Parts \cite{korsch_classification-specific_2019} & ResNet-50 & 88.5 \\
        MGE-CNN \cite{zhang_learning_2019} & ResNet-101 & 88.6 \\
        CAP \cite{behera_context-aware_2021} & Xception & 91.0 \\
        \midrule
        ViT B-16 \cite{dosovitskiy_image_2020} & ViT B-16 & 89.9 \\
        DeiT B-16 \cite{touvron_training_2021} & DeiT B-16 & 87.3 \\
        TPSKG \cite{liu_transformer_2022} & ViT B-16 & 90.1 \\
        TransFG \cite{he_transfg_2022} & ViT B-16 & 90.8 \\
        IELT \cite{xu_fine-grained_2023} & ViT B-16 & 90.8 \\
        Dual-TR \cite{ji_dual_2023} & ViT B-16 & \underline{91.3} \\
        R2-Trans \cite{ye_r2-trans_2024} & ViT B-16 & 90.2 \\
        DeiT-NeT \cite{kim_neural_2024} & DeiT B-16 & 88.4 \\
        GLSim (Ours) & ViT B-16 & \textbf{92.0} \\
        \bottomrule
    \end{tabularx}

    \caption{Recognition accuracy for state-of-the-art models on NABirds \cite{van_horn_building_2015} dataset with image size 448x448.}
    \label{table_results_acc_nabirds}

\end{table}

\subsubsection{Comparison on iNat17}

In the challenging iNat17 \cite{van_horn_inaturalist_2018}, our method attains an improvement of 1.5\% compared to the second best performing method, DeiT-NeT \cite{kim_neural_2024} which employs a ViT trained with the DeiT \cite{touvron_training_2021} pretraining recipe. Among models using the ViT B-16 pretraining recipe, our model outperforms the next best, TransFG \cite{he_transfg_2022} by 3.8\% absolute accuracy, while the baseline ViT by 6.8\%. This increase in relative improvement on the iNat17 dataset shows the promise behind our proposed methodology for large-scale FGIR tasks. 

\begin{table}[htb]
    \centering
    \begin{tabularx}{\linewidth}{l l Y}
        \toprule
        Method & Backbone & Top-1 (\%) \\
        \midrule
        ResNet-152 \cite{he_deep_2015} & ResNet-152 & 59.0 \\
        SpineNet-143 \cite{du_spinenet_2020} & SpineNet-143 & 63.6 \\
        IncResNetV2 SE \cite{van_horn_inaturalist_2018} & IncResNetV2 & 67.3 \\
        SSN \cite{ferrari_learning_2018} & ResNet-101 & 65.2 \\
        TASN \cite{zheng_looking_2019} & ResNet-101 & 68.2 \\
        \midrule
        ViT B-16 \cite{dosovitskiy_image_2020} & ViT B-16 & 68.7 \\
        DeiT B-16 \cite{touvron_training_2021} & DeiT B-16 & 72.8 \\
        RAMS-Trans \cite{hu_rams-trans_2021} & ViT B-16 & 68.5 \\
        AFTrans \cite{zhang_free_2022} & ViT B-16 & 68.9 \\
        SIM-Trans \cite{sun_sim-trans_2022} & ViT B-16 & 69.9 \\
        TransFG \cite{he_transfg_2022} & ViT B-16 & 71.7 \\
        Dual-TR \cite{ji_dual_2023} & ViT B-16 & 71.5 \\
        DeiT-NeT \cite{kim_neural_2024} & DeiT B-16 & \underline{74.0} \\
        GLSim (Ours) & ViT B-16 & \textbf{75.5} \\
        \bottomrule
    \end{tabularx}

    \caption{Recognition accuracy for state-of-the-art models on iNat17 \cite{van_horn_inaturalist_2018} dataset with image size 304x304.}
    \label{table_results_acc_inat17}

\end{table}

\subsubsection{Comparison Using Image Size 224x224}

\begin{table*}[htb]
    \centering

    \small{
    \begin{tabularx}{\linewidth}{l Y Y Y Y Y Y}
        \toprule
        \multirow{2}{*}{Dataset} & \multicolumn{2}{c}{CNN-Based} & \multicolumn{4}{c}{ViT-Based}\\
        \cmidrule(lr){2-3} \cmidrule(lr){4-7}
        {} & RN-101 \cite{he_deep_2015} & CAL \cite{rao_counterfactual_2021} & ViT B-16  \cite{dosovitskiy_image_2020} & TransFG \cite{he_transfg_2022} & FFVT \cite{wang_feature_2021} & Ours\\
        \midrule
        CUB \cite{wah_caltech-ucsd_nodate} & $82.71\pm{0.06}$ & $88.44\pm{0.25}$ & $90.11\pm{0.17}$ & $\underline{90.49\pm{0.06}}$ & $90.05\pm{0.04}$ & $\mathbf{90.93\pm{0.24}}$\\
        DAFB \cite{branwen_danbooru2021_2015} & $91.20\pm{0.08}$ & $\mathbf{95.15\pm{0.01}}$ & $91.78\pm{0.06}$ & $92.87\pm{0.01}$ & $91.49\pm{0.01}$ & $\underline{93.36\pm{0.02}}$\\
        Dogs \cite{khosla_novel_2011} & $91.10\pm{0.18}$ & $89.78\pm{0.10}$ & $91.36\pm{0.05}$ & $90.68\pm{0.14}$ & $\underline{91.49\pm{0.01}}$ & $\mathbf{92.01\pm{0.09}}$\\
        Flowers \cite{nilsback_automated_2008} & $92.36\pm{0.24}$ & $98.14\pm{0.04}$ & $\underline{99.45\pm{0.03}}$ & $99.44\pm{0.03}$ & $99.32\pm{0.02}$ & $\mathbf{99.48\pm{0.02}}$\\
        Food \cite{bossard_food-101_2014} & $87.45\pm{0.21}$ & $89.71\pm{0.12}$ & $92.43\pm{0.04}$ & $\underline{92.76\pm{0.05}}$ & $91.89\pm{0.01}$ & $\mathbf{92.97\pm{0.04}}$\\
        iNat17 \cite{van_horn_inaturalist_2018} & $63.37\pm{0.16}$ & $69.29\pm{0.03}$ & $69.38\pm{0.04}$ & $\underline{72.34\pm{0.12}}$ & $69.09\pm{0.04}$ & $\mathbf{73.50\pm{0.06}}$\\
        Moe \cite{noauthor_tagged_nodate} & $94.22\pm{0.06}$ & $\mathbf{97.55\pm{0.00}}$ & $96.47\pm{0.11}$ & $96.59\pm{0.20}$ & $\underline{96.67\pm{0.15}}$ & $\underline{96.67\pm{0.16}}$\\
        NABirds \cite{van_horn_building_2015} & $81.00\pm{0.26}$ & $88.33\pm{0.49}$ & $87.36\pm{0.09}$ & $\underline{88.76\pm{0.08}}$ & $86.63\pm{0.08}$ & $\mathbf{89.11\pm{0.16}}$\\
        Pets \cite{parkhi_cats_2012} & $93.20\pm{0.19}$ & $93.89\pm{0.23}$ & $94.10\pm{0.07}$ & $93.93\pm{0.14}$ & $\underline{94.34\pm{0.11}}$ & $\mathbf{95.01\pm{0.18}}$\\
        VegFru \cite{hou_vegfru_2017} & $89.34\pm{0.05}$ & $90.33\pm{4.28}$ & $95.77\pm{0.05}$ & $\underline{95.91\pm{0.04}}$ & $95.58\pm{0.02}$ & $\mathbf{96.14\pm{0.04}}$\\
        \midrule
        Average & $86.59\pm{9.25}$ & $90.06\pm{8.16}$ & $90.84\pm{8.29}$ & $\underline{91.38\pm{7.41}}$ & $90.56\pm{8.39}$ & $\mathbf{91.92\pm{7.14}}$\\
        \bottomrule
        \end{tabularx}
    }

    \caption{Recognition accuracy for state-of-the-art models with image size 224x224.}
    \label{table_results_acc_is_224}
\end{table*}

Previous research suggested the applicability of FGIR methods is task dependent \cite{ye_image_2024}. Therefore, to obtain a better understanding of how our method performs across FGIR tasks we conduct a comparison across 10 FGIR datasets using image size 224x224 in \Cref{table_results_acc_is_224}. Our proposed method, GLSim, obtains the highest accuracy in 8 datasets (CUB, Dogs, Flowers, Food, iNat17, NABirds, Pets, VegFru) and the second best in 2 others (DAFB, Moe). On average our method obtains the highest classification accuracy, reducing the relative classification error by 10.15\% compared to the baseline ViT. These results demonstrate the robustness of GLSim across a wide variety of FGIR domains.


\subsection{Qualitative evaluation}
\label{ssec_qualitative}

\begin{figure}[htb]
    \centering
    
    \includegraphics[width=1.0\linewidth]{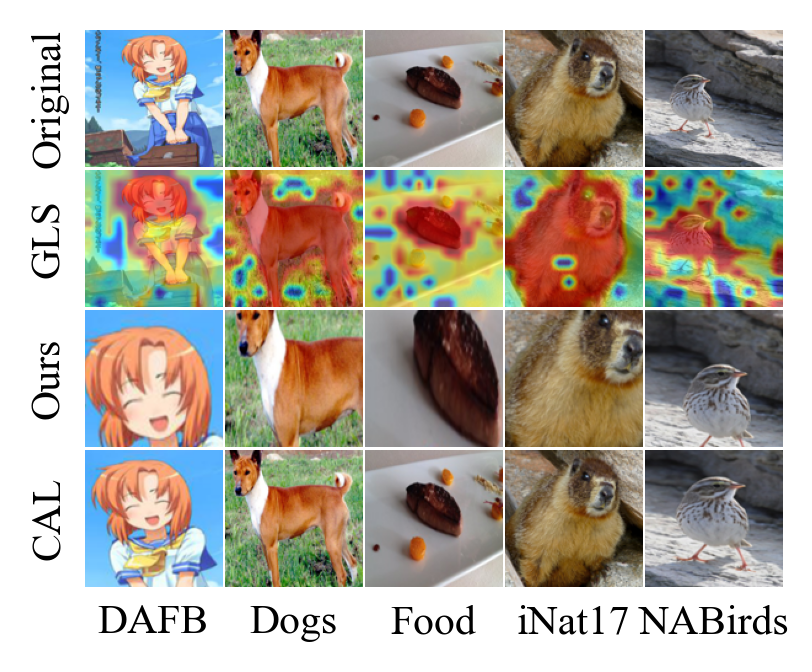}

    \caption{Visualization of samples from various fine-grained datasets. First row shows the original images. Second and third rows show the heatmap for the proposed global-local similarity (GLS) metric and crops obtained based on it. Fourth row shows crops for CAL \cite{rao_counterfactual_2021}.}
    \label{figure_crops_sample}

\end{figure}

We evaluate the relation of the proposed global-local similarity to the quality of our crops and also compare them to crops from CAL by visualizing samples from various fine-grained datasets in \Cref{figure_crops_sample}. In general, our method adeptly capture the objects of interest present within the corresponding images, thereby mitigating the impact of background noise. This, in turns, facilitates the extraction of finer details during the second forward pass.

When comparing our cropping method to CAL on the DAFB dataset, we observe that our approach yields more zoomed-in crops than those produced by CAL. This tighter cropping may inadvertently exclude critical details necessary for effective discrimination in this task. Additionally, CAL's use of multiple crops could account for the observed accuracy gap between our method and CAL.


\subsection{Computational Cost Analysis}
\label{ssec_computational_cost}

As the accuracy of FGIR models reaches a certain threshold, the computational cost of deploying them becomes a critical factor. We compare the trade-off between accuracy and the throughput (in images per second) at inference time when batching the images, and the associated VRAM required for computing this batched inference.

\begin{figure}[htb]
    \centering
    
    \includegraphics[width=1.0\linewidth]{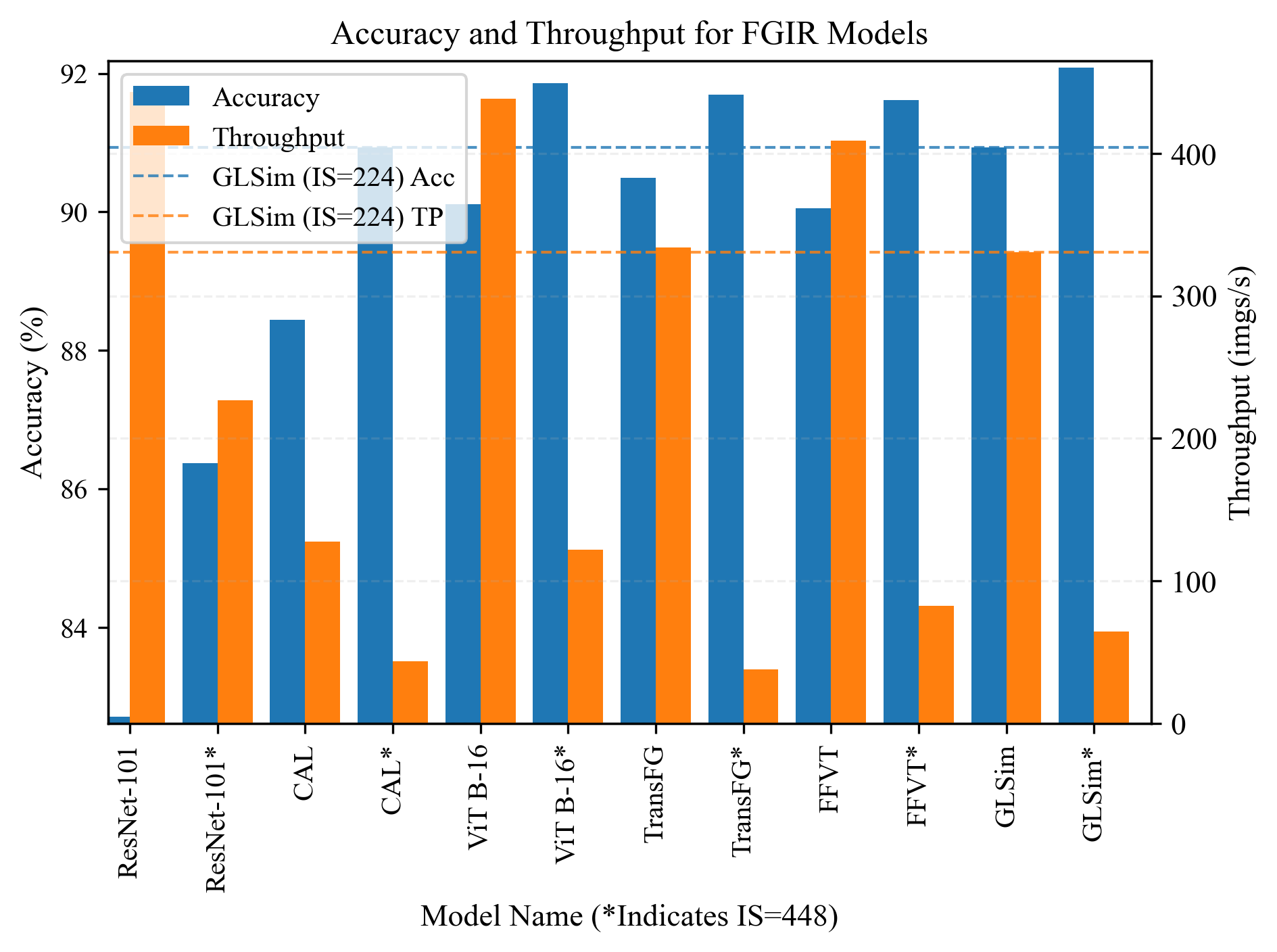}

    \caption{Accuracy and throughput plots for the evaluated models on CUB. \textit{Model}$^*$ represents results for image size 448x448. We highlight the accuracy and throughput for our proposed GLSim method with image size 224x224.}
    \label{figure_acc_tp}
\end{figure}

We show the results on \Cref{figure_acc_tp} on accuracy vs inference throughput as for many users the throughput is a limiting factor when deploying models with real-time requirements. From this figure, we can observe that GLSim with image size 224 obtains a competitive accuracy, that is only surpassed by ViT, TransFG, FFVT and GLSim with image size 448, albeit slightly. However, the throughput for GLSim with image size 224 is 2.70x, 8.73x, 4.00x, and 5.11x times higher compared to the aforementioned models with image size 448. 

With regards to VRAM, we highlight the low memory requirements of GLSim  with image size 224 as it is the only model to require less than 1 GB of VRAM during inference. This is because unlike TransFG and FFVT it does not require the storage and processing of intermediate features. GLSim requires 5.45x and 5.29x less VRAM compared to TransFG, and 2.32x and 1.65x less VRAM compared to FFVT, for image sizes 224 and 448, respectively.

Furthermore, when compared to CAL, which yields the best results in DAFB and Moe, our proposed method has higher throughput, and requires substantially less VRAM during inference. Specifically, our method has 2.59x and 1.48x higher throughput, and requires 9.26x and 8.11x less VRAM, for image size 224 and 448, respectively.

Overall, these results suggest that our proposed method achieves high accuracy while keeping computational cost low, making it a practical and efficient solution for deploying FGIR models in real-world applications.


\subsection{Ablation on Proposed Components}
\label{ssec_ablations}

We present a breakdown of the individual contributions of the proposed components of our system in \Cref{table_ablation_components}. The first row outlines the performance of the baseline ViT B-16 model, while the second row reflects the outcomes of a modified ViT with an extra encoder block that only processes the CLS token. The third row describes a scenario where we select and encode image crops but do not explicitly combine high-level features. Instead, we choose the prediction with the highest confidence between the original image and the cropped image. Finally, the fourth row represents our complete system, GLSim.

The proposed Aggregator and GLS Cropping modules reduce the relative classification error by 4.41\% and 4.38\%, respectively. By incorporating both modules the error is further reduced by an additional 4.12\% compared to cropping without explicit high-level feature aggregation. This highlights the importance of the proposed modules to the overall effectiveness of our system.

\begin{table}[htb]
    \centering

    \small{
    \begin{tabularx}{\linewidth}{Y Y Y}
        \toprule
        GLS & Aggregator & Accuracy (\%)\\
        \midrule
        - & - & $90.11\pm{0.17}$\\
        \midrule
        - & $\checkmark$ & $90.55\pm{0.06}$\\
        \midrule
        $\checkmark$ & - & $90.54\pm{0.16}$\\
        \midrule
        $\checkmark$ & $\checkmark$ & $\mathbf{90.93\pm{0.24}}$\\
        \bottomrule
        \end{tabularx}
    }

    \caption{Ablation on proposed components on CUB.}
    \label{table_ablation_components}
\end{table}

\section{Future Work: GLS For Visualization and Downstream Tasks}
\label{sec_discussion and future prospects}


\begin{figure}[htb]
    \centering
    
    \includegraphics[width=1.0\linewidth]{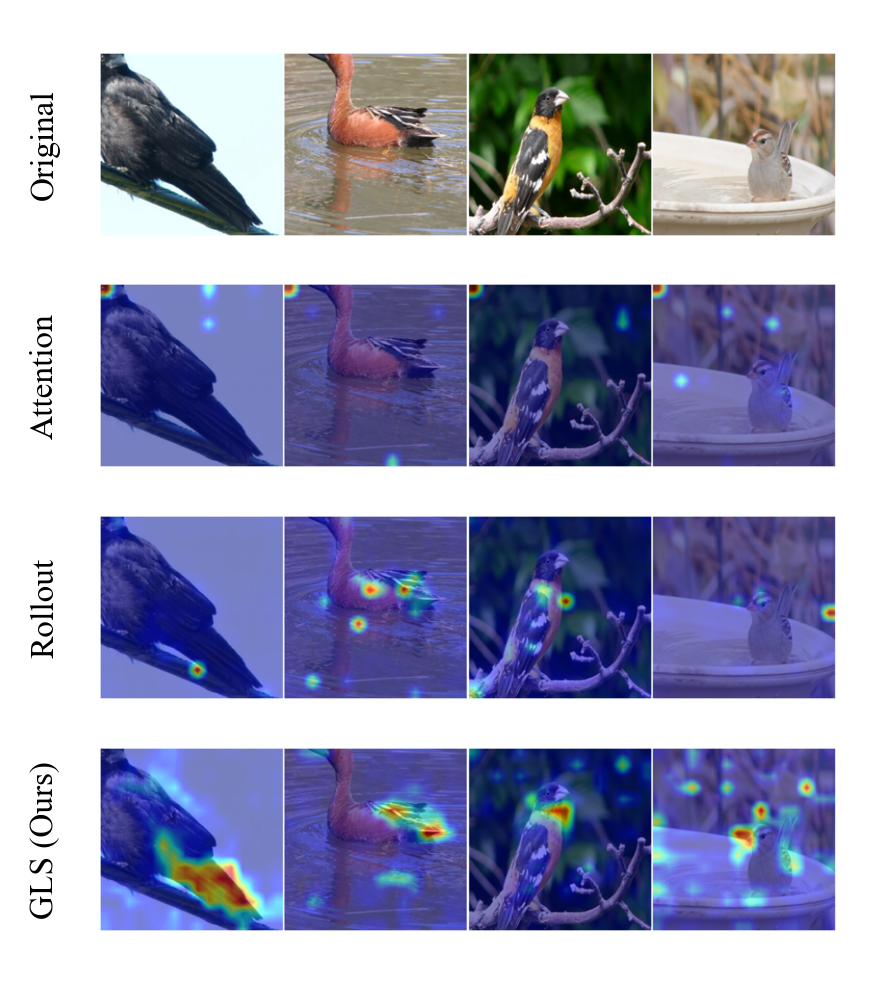}

    \caption{Visualization of attention-based visualization mechanisms and our proposed GLS for DINOv2 B-14 \cite{oquab_dinov2_2023} on NABirds.}
    \label{figure_dfsm_dinov2}
\end{figure}


The proposed GLS metric can be used as a visualization tool to highlight which local regions of the image have high similarity to the global representation from the CLS token to interpret classification predictions. Furthermore, GLS shows even better discrimination performance when combined with state-of-the-art pretrained backbones such as DINOv2 \cite{oquab_dinov2_2023} as shown in \Cref{figure_dfsm_dinov2}. This could allow deploying these models in a variety of downstream tasks such as fine-grained weakly supervised localization and semantic segmentation \cite{sohn_fine-grained_2021, metzger_fine-grained_2021}, fine-grained object recognition based on remote sensing imagery \cite{guo_fine-grained_2023, wang_efficient_2023, sun_fair1m_2022} where resources are constrained. We remark how the computational cost of our proposed GLS can range from \textbf{648x to 271,299x less} compared to matrix-multiplication based attention aggregation mechanisms \cite{conde_exploring_2021, hu_rams-trans_2021, zhu_dual_2022, he_transfg_2022, liu_transformer_2022}, for a B-14 backbone with image size ranging from 224 to 1024.

\section{Conclusion}
\label{sec_conclusion}

This paper proposes GLS, an efficient and effective alternative to attention scores in vision transformers, for the purpose of discriminative region identification to enhance fine-grained image recognition. GLS compares the similarity between global and local representations of an image. Based on this we propose a system GLSim which extracts discriminative crops and combines high-level features from both the original and cropped image using an aggregator module for robust predictions. Extensive evaluations across various datasets demonstrate the robustness of our method.

{\small
\bibliographystyle{ieee_fullname}
\bibliography{main}
}

\clearpage
\appendix

\section*{Appendix for Global-Local Similarity
for Efficient Fine-Grained Image Recognition with Vision Transformers}

\section{Extended Comparison of ViTs for FGIR}
\label{sec_appendix_extended_comparison_vits_fgir}

\begin{table*}[!htbp]
    \centering

    \small{
    \begin{tabularx}{\linewidth}{X d d X d X X}
        \toprule
        Model & PO & IF & DFSM & Crops & FA & Complexity\\
        \midrule
        TransFG \cite{he_transfg_2022} & $\checkmark$ & $\checkmark$ & PSM & - & Transformer & $L\cdot H \cdot N^3$\\
        FFVT \cite{wang_feature_2021} & - & $\checkmark$ & MAWS & - & Transformer & $L\cdot H\cdot N$\\
        RAMS-Trans \cite{hu_rams-trans_2021} & - & $\checkmark$ & Rollout & $\checkmark$ & - & $L\cdot N^3$\\
        AF-Trans \cite{zhang_free_2022} & $\checkmark$ & $\checkmark$ & SACM & $\checkmark$ & - & $L\cdot H \cdot N^2$\\
        DCAL \cite{zhu_dual_2022} & - & $\checkmark$ & Rollout & - & X-Attention & $L\cdot N^3$\\
        Ours & - & - & GLS & $\checkmark$ & Transformer & $N\cdot D$\\
        \bottomrule
    \end{tabularx}
    }

    \caption{Summary of differences between ViTs for FGIR.}
    \label{table_model_comparison_all}
\end{table*}

\begin{table*}[!htbp]
    \centering

    \small{
    \begin{tabularx}{\linewidth}{X X X X X X}
        \toprule
        Model & TransFG \cite{he_transfg_2022} & Various \cite{hu_rams-trans_2021, zhu_dual_2022, liu_transformer_2022} & FFVT \cite{wang_feature_2021} & AF-Trans \cite{hu_rams-trans_2021} & Ours \\
        \midrule
        DFSM & PSM & Rollout & MAWS & SACM & GLS \\
        Complexity & $L\cdot H \cdot N^3$ & $L\cdot N^3$ & $L\cdot H\cdot N$ & $L\cdot H \cdot N^2$ & $N\cdot D$ \\
        \midrule

        FLOPs (IS=224) & $2.7\times 10^4$ & $3.8\times 10^2$ & $1.7\times 10^-1$ & $3.1\times 10^1$ & $5.9\times 10^-1$ \\
        \% of Backbone & \textbf{117.51} & 1.6455 & 0.0007 & 0.1344 & 0.0025 \\
        \midrule
        FLOPs (IS=448) & $1.7\times 10^6$ & $2.4\times 10^4$ & $6.9\times 10^-1$ & $5.0\times 10^2$ & $2.4\times 10^0$ \\
        \% of Backbone & \textbf{1624.4} & 22.212 & 0.0006 & 0.4649 & 0.0022 \\
        \midrule
        FLOPs (IS=768) & $4.6\times 10^7$ & $5.5\times 10^5$ & $2.0\times 10^0$ & $4.4\times 10^3$ & $6.7\times 10^0$ \\
        \% of Backbone & \textbf{11362} & \textbf{134.39} & 0.0005 & 1.09 & 0.0017 \\
        \midrule
        FLOPs (IS=1024) & $2.7\times 10^8$ & $3.3\times 10^6$ & $3.6\times 10^0$ & $1.4\times 10^4$ & $1.2\times 10^1$ \\
        \% of Backbone & \textbf{27554} & \textbf{339.94} & 0.0004 & 1.4676 & 0.0013 \\
        \bottomrule
    \end{tabularx}
    }

    \caption{Summary of computational cost for discriminative feature selection mechanisms (DFSM) used by ViTs in FGIR and their ratio of FLOPs compared to the backbone for ViT B-14. We highlight in \textbf{bold} values that exceed 50\% of the FLOPs of the backbone.}
    \label{table_model_comparison_b14}
\end{table*}


\begin{table*}[!htbp]
    \centering

    \small{
    \begin{tabularx}{\linewidth}{X X X X X X}
        \toprule
        Model & TransFG \cite{he_transfg_2022} & Various \cite{hu_rams-trans_2021, zhu_dual_2022, liu_transformer_2022} & FFVT \cite{wang_feature_2021} & AF-Trans \cite{hu_rams-trans_2021} & Ours \\
        \midrule
        DFSM & PSM & Rollout & MAWS & SACM & GLS \\
        Complexity & $L\cdot H \cdot N^3$ & $L\cdot N^3$ & $L\cdot H\cdot N$ & $L\cdot H \cdot N^2$ & $N\cdot D$ \\
        \midrule

        FLOPs (IS=224) & $2.1\times 10^3$ & $1.7\times 10^2$ & $3.5\times 10^-2$ & $2.6\times 10^0$ & $1.1\times 10^-1$ \\
        \% of Backbone & \textbf{160.36} & 13.228 & 0.0027 & 0.2062 & 0.0089 \\
        \midrule
        FLOPs (IS=448) & $1.5\times 10^5$ & $1.1\times 10^4$ & $1.4\times 10^-1$ & $4.7\times 10^1$ & $4.5\times 10^-1$ \\
        \% of Backbone & \textbf{2130.8} & \textbf{147.35} & 0.0019 & 0.6485 & 0.0063 \\
        \midrule
        FLOPs (IS=768) & $3.8\times 10^6$ & $2.7\times 10^5$ & $4.1\times 10^-1$ & $3.9\times 10^2$ & $1.3\times 10^0$ \\
        \% of Backbone & \textbf{10037} & \textbf{720.95} & 0.0011 & 1.0537 & 0.0036 \\
        \midrule
        FLOPs (IS=1024) & $2.3\times 10^7$ & $1.5\times 10^6$ & $7.2\times 10^-1$ & $1.3\times 10^3$ & $2.4\times 10^0$ \\
        \% of Backbone & \textbf{22568} & \textbf{1509.0} & 0.0007 & 1.3015 & 0.0024 \\
        \bottomrule
    \end{tabularx}
    }

    \caption{Summary of computational cost for discriminative feature selection mechanisms (DFSM) used by ViTs in FGIR and their ratio of FLOPs compared to the backbone for ViT T-16. We highlight in \textbf{bold} values that exceed 50\% of the FLOPs of the backbone.}
    \label{table_model_comparison_t16}
\end{table*}

We summarize the differences between various ViT models proposed for FGIR in \Cref{table_model_comparison_all}. The following aspects are considered: (i) Patch Overlap (PO), which indicates whether the ViT's patchifier convolution has overlapping stride; (ii) Intermediate Features (IF), which refers to whether the model uses intermediate values (features or attention scores); (iii) Discriminative Feature Selection Mechanism (DFSM), which describes how the model selects discriminative features; (iv) Crops, which indicates whether the model crops the image for a second forward pass; (v) Feature Aggregation (FA), which denotes the modules used for performing discriminative feature refinement or aggregation; and (vii) Complexity. The `$\checkmark$' represents the model uses the feature, while `-' represents it was not.

We observe that 1) many of the proposed modifications to the original ViT are modular and can be substituted or combined and 2) the most significant difference between these methods lies in their DFSM. For this reason, we focus on this aspect in Table \ref{table_model_comparison_b16} in the main text, and Tables \ref{table_model_comparison_b14} and \ref{table_model_comparison_t16} in the Appendix.

We note how the cubic complexity of matrix-multiplication based attention aggregation mechanisms is pervasive across architectures with different patch size (see \Cref{table_model_comparison_b14}) and model capacity (\Cref{table_model_comparison_t16}). Furthermore, the percentage of computation required by these mechanisms increases as the patch size or model capacity decreases. This can hinder the applicability of attention rollout and similar mechanisms in resource constrained environments.

\section{Pseudocode for Our Method}
\label{sec_appendix_pseudocode}

\begin{figure}[htb]
    \centering
    
    \includegraphics[width=0.9\linewidth]{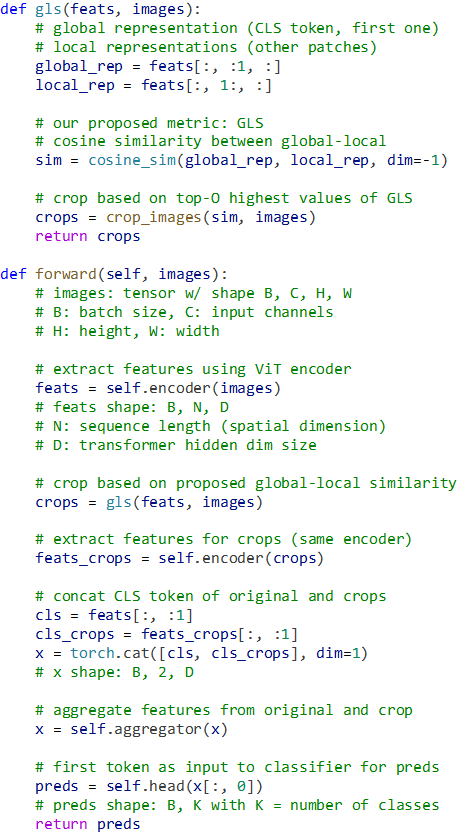}

    \caption{Pytorch-like pseucodode for the forward pass of GLSim.}
    \label{figure_pseudocode}
\end{figure}

We include Pytorch-like pseudocode to facilitate the understanding of our method, noting the tensor dimensions at the different steps of the forward pass, in \Cref{figure_pseudocode}.


\section{Extended Experimental Setup}
\label{sec_appendix_extended_experimental_setup}

\begin{table}[htbp]
    \centering

    \small{
    \begin{tabularx}{\linewidth}{l X X X X X}
        \toprule
        Dataset & \ $T$ & \# Train & \# Test & $\overline{S_1}$ & $\overline{S_2}$\\
        \midrule
        CUB \cite{wah_caltech-ucsd_nodate} & 200 & 5,994 & 5,794 & 468 & 386\\
        \midrule
        DAFB \cite{branwen_danbooru2021_2015, rios_anime_2022} & 3,263 & 368,997 & 94,440 & 268 & 340\\
        \midrule
        Dogs \cite{khosla_novel_2011} & 120 & 12,000 & 8,580 & 443 & 386\\
        \midrule
        Flowers \cite{nilsback_automated_2008} & 102 & 2,040 & 6,149 & 623 & 536\\
        \midrule
        Food \cite{bossard_food-101_2014} & 101 & 75,750 & 25,250 & 496 & 475\\
        \midrule
        iNat17 \cite{van_horn_inaturalist_2018} & 5,089 & 579,184 & 95,986 & 746 & 645\\
        \midrule
        Moe \cite{noauthor_tagged_nodate} & 173 & 11,454 & 2,943 & 159 & 146\\
        \midrule
        NABirds \cite{van_horn_building_2015} & 600 & 23,929 & 24,633 & 901 & 713\\
        \midrule
        Pets \cite{parkhi_cats_2012} & 37 & 3,680 & 3,669 & 431 & 383\\
        \midrule
        VegFru \cite{hou_vegfru_2017} & 292 & 43,800 & 116,391 & 347 & 287\\
        \bottomrule
        \end{tabularx}
    }

    \caption{Statistics for evaluated datasets. $T$, $\overline{S_1}$ and $\overline{S_2}$ Represent number of classes, average width and height, respectively.}
    \label{table_dataset_stats}
\end{table}

\begin{table}[htbp]
    \centering

    \small{
    \begin{tabularx}{\columnwidth}{d X}  
        \toprule
        Setting & Value\\
        \midrule
        Models & \{ResNet-101 \cite{he_deep_2015}, CAL \cite{rao_counterfactual_2021}, ViT B-16 \cite{dosovitskiy_image_2020}, TransFG \cite{he_transfg_2022}, FFVT \cite{wang_feature_2021}, GLSim\}\\
        \midrule
        Optimizer & SGD\\
        \midrule
        LR range & \{0.03, 0.01, 0.003, 0.001\}\\
        \midrule
        Batch size & 8\\
        \midrule
        Weight decay & $0^*$\\
        \midrule
        Epochs & $50^{*}$\\
        \midrule
        LR Schedule & Cosine with Warmup for 500* Steps\\
        \midrule
        Image size & \{224, 448\}\\
        \midrule
        Preprocess & {Normalize then square resize followed by random (train) or center (inference) crop}\\
        \midrule
        Augmentations & Random Horizontal Flip, Random Erasing \cite{zhong_random_2020}, Trivial Augment \cite{muller_trivialaugment_2021}\\
        \midrule
        Regularization & Label Smoothing \cite{szegedy_rethinking_2016}, Stochastic Depth \cite{huang_deep_2016}\\
        \bottomrule
        \end{tabularx}
    }

    \caption{Training (and Inference) Hyperparameter Settings. *With exceptions for CAL. Details are described in \Cref{sec_appendix_extended_experimental_setup}.}
    \label{table_experiment_settings}
\end{table}

Previous research suggested the applicability of FGIR methods is task dependent \cite{ye_image_2024}. Therefore, to obtain a better understanding of how our method performs across FGIR tasks we conduct a comparison across 10 diverse FGIR datasets. The statistics for the evaluated datasets are shown in \Cref{table_dataset_stats}.

The training and test settings employed in our experiments are summarized in Table \ref{table_experiment_settings}. For each model and dataset pair, we first perform a learning rate (LR) search with a training and validation set created by partitioning 80\% and 20\% of the total training data (train-val), respectively. However, for datasets with more than 50,000 images in the train-val set (DAFB, Food, iNat17), we set the partition ratio to 20\% and 80\%. We subsequently train the models on the entire training-validation set, using three different seeds (1, 10, 100) each, except for the larger datasets mentioned earlier, where we only employ two seeds (1, 10). The models are trained for 50 epochs, except for the CAL model, which is trained for 160 epochs as per the original author's recommendation, except for DAFB, Food, and iNat17, where we train for 50 epochs.

We employ the SGD optimizer with batch size of 8, weight decay of 0 (except for CAL, where it is set to $5\times 10^{-4}$), and cosine annealing \cite{loshchilov_sgdr_2017} with a warmup of 500 steps (1,500 for CAL) for learning rate scheduling.

Regarding the preprocessing of images, we first normalize them and then resize them to a square of size $1.34S_1\times 1.34S_2$. We then perform a random crop during training and a center crop during inference. 

In addition to the application of random horizontal flipping during training, we further incorporate a stronger set of augmentation and regularization (AugReg) techniques that have been shown to improve the generalization capabilities of models and mitigate overfitting. Our choice of AugReg is motivated by recent advances in training strategies for coarse image recognition \cite{touvron_training_2021, touvron_deit_2022}. Specifically, random erasing \cite{zhong_random_2020} and trivial augment \cite{muller_trivialaugment_2021} are employed for the purpose of data augmentation. For regularization, we employ label smoothing \cite{szegedy_rethinking_2016} and stochastic depth \cite{huang_deep_2016}, albeit the latter only for ViT-based models.

\section{Extended Results and Discussion}
\label{sec_appendix_extended_results}

\subsection{Extended Analysis on Computational Cost}


\begin{figure}[htb]
    \centering
    
    \includegraphics[width=1.0\linewidth]{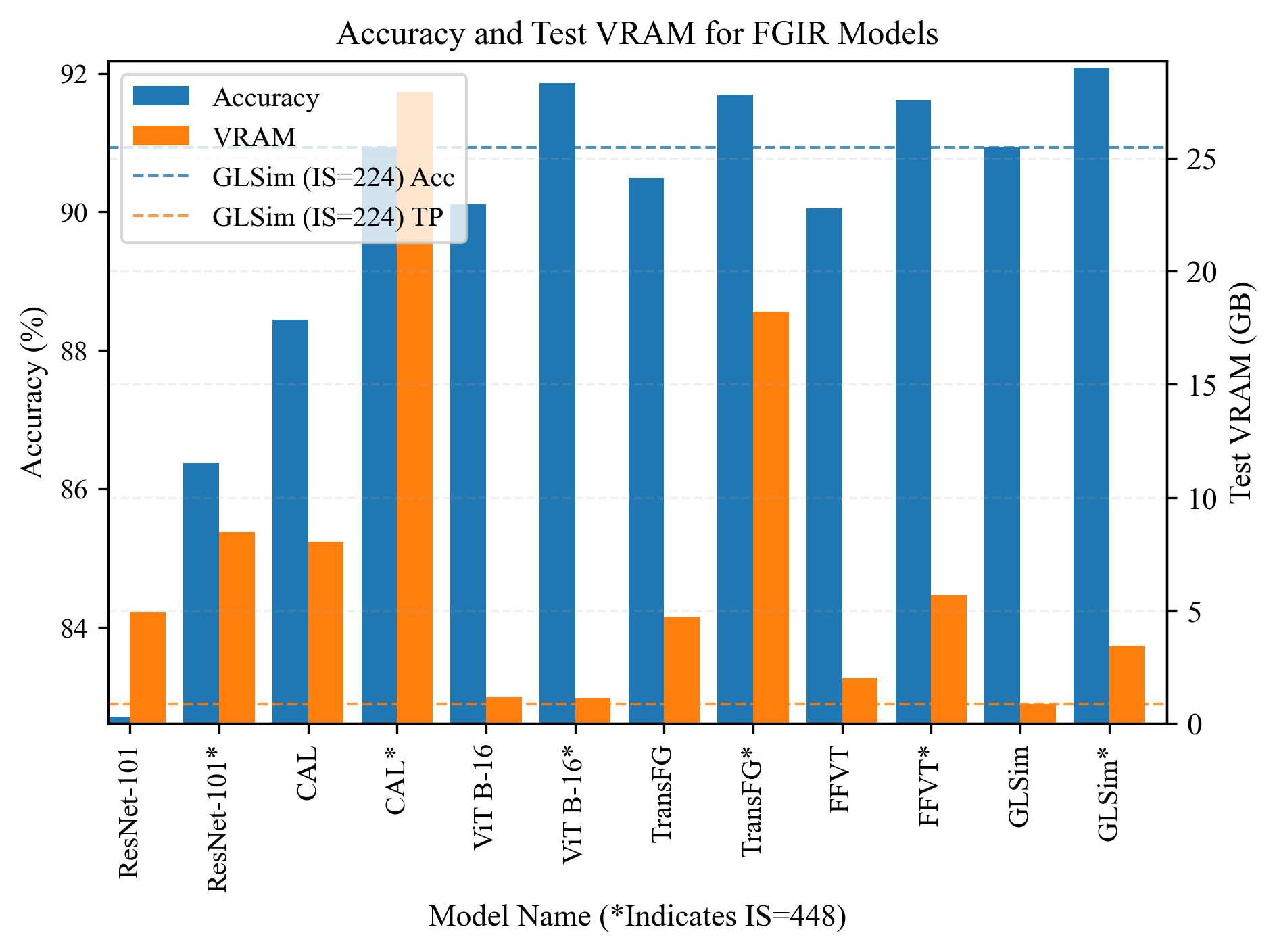}

    \caption{Accuracy and VRAM plots for the evaluated models on CUB. \textit{Model}$^*$ represents results for image size 448x448.}
    \label{figure_acc_vram}
\end{figure}


We compare the trade-off between accuracy and the throughput (in images per second) at inference time when batching the images, and the associated VRAM (in Gigabytes) required for computing this batched inference. For the batched throughput we first compare the throughput at different batch sizes, from 1 to 256 for all these model-image size pairs, and select the batch size which led to the highest throughput (and corresponding VRAM requirements).

In \Cref{figure_acc_tp} of the main text we show the results on accuracy vs inference throughput. We additionally show the results on \Cref{figure_acc_vram} on accuracy vs inference memory requirements as the VRAM requirement is also an important decision when deploying models. Discussion is included in \Cref{ssec_computational_cost} of the main text.

\subsection{Effects of Augmentation and Regularization on FGIR}
\label{ssub_effects_augs}

While the impact of generic Augmentations and Regularizations (AugReg) on coarse image recognition has been widely studied, its influence on FGIR has received less attention. In general, newer coarse recognition backbones are trained using strong augmentations and regularization techniques, such as Random Erasing (RE) \cite{zhong_random_2020}, AutoAugment (AA) \cite{cubuk_autoaugment_2019}, TrivialAugment (TA) \cite{muller_trivialaugment_2021} Label Smoothing (LS) \cite{szegedy_rethinking_2016}, and Stochastic Depth (SD) \cite{huang_deep_2016}. However, in FGIR, works typically only incorporate random cropping (RC) and random horizontal flipping (RF), with some newer works \cite{he_transfg_2022, wang_feature_2021} also using LS.

The reasoning behind this is that strong augmentations may obfuscate the subtle differences required to distinguish between fine-grained categories. As a result, a line of FGIR research has emerged, focused on designing data-aware AugReg techniques. One prominent work in this area is WS-DAN \cite{hu_see_2019}, which proposes using attention maps to guide cropping and masking of a given image during training. These two techniques are employed in alternation throughout the training process to both augment the input images the network processes and regularize the network, preventing overfitting. Furthermore, during inference multiple crops are obtained to maximize recognition performance. CAL \cite{rao_counterfactual_2021} builds on WS-DAN by using counterfactual causality to measure the attention quality and guide this process. However, as seen in \Cref{figure_acc_tp} in the main text this process incurs considerable computational cost during inference.

\begin{table*}[htb]
    \centering

    \small{

    \begin{tabularx}{\linewidth}{l Y Y Y Y}
        \toprule
        \multirow{2}{*}{Model} & Minimal & Weak & Medium & Strong\\
        {} & (RC + RF) & (++LS) & (++SD) & (++RE + TA)\\
        \midrule
        ResNet-101 & $80.85\pm{0.62}$ & $\underline{81.03\pm{0.29}}$ & $80.89\pm{0.55}$ & $\mathbf{82.71\pm{0.06}}$ \\
        CAL & $86.39\pm{0.89}$ & $87.01\pm{0.12}$ & $\underline{87.36\pm{0.28}}$ & $\mathbf{88.44\pm{0.25}}$ \\
        \midrule
        ViT B-16  & $88.27\pm{0.2}$ & $88.53\pm{0.27}$ & $\underline{89.5\pm{0.21}}$ & $\mathbf{90.11\pm{0.17}}$\\
        TransFG  & $89.06\pm{0.23}$ & $88.94\pm{0.16}$ & $\underline{89.93\pm{0.24}}$ & $\mathbf{90.49\pm{0.06}}$\\
        FFVT & $88.83\pm{0.04}$ & $88.78\pm{0.09}$ & $\mathbf{90.15\pm{0.2}}$ & $\underline{90.05\pm{0.04}}$ \\
        GLSim  & $89.78\pm{0.13}$ & $89.72\pm{0.38}$ & $\underline{90.49\pm{0.26}}$ & $\mathbf{90.93\pm{0.24}}$\\
        \midrule
        \midrule
        ResNet-101$^*$  & $84.72\pm{0.25}$ & $85.16\pm{0.39}$ & $\underline{85.25\pm{0.45}}$ & $\mathbf{86.37\pm{0.22}}$ \\
        CAL$^*$  & $89.39\pm{0.11}$ & $\underline{88.9\pm{0.48}}$ & $87.61\pm{1.53}$ & $\mathbf{90.94\pm{0.21}}$ \\
        \midrule
        ViT B-16$^*$  & $90.13\pm{0.13}$ & $90.51\pm{0.17}$ & $\underline{91.44\pm{0.14}}$ & $\mathbf{91.86\pm{0.01}}$ \\
        TransFG$^*$  & $90.6\pm{0.11}$ & $90.31\pm{0.07}$ & $\underline{91.25\pm{0.02}}$ & $\mathbf{91.69\pm{0.15}}$ \\
        FFVT$^*$  & $90.1\pm{0.22}$ & $90.05\pm{0.08}$ & $\mathbf{91.69\pm{0.1}}$ & $\underline{91.61\pm{0.04}}$ \\
        GLSim$^*$  & $91.06\pm{0.08}$ & $91.1\pm{0.19}$ & $\underline{91.89\pm{0.14}}$ & $\mathbf{92.09\pm{0.01}}$ \\
        \bottomrule
    \end{tabularx}
    }

    \caption{Effects of Augmentations and Regularization (AugReg) on Fine-Grained Recognition Accuracy for CUB Dataset. ++ Represents All Previous AugReg are Used. RC, RF, LS, SD, RE, and TA stand for Random Crop, Random Horizontal Flip, Label Smoothing, Stochastic Depth, Random Erase, and Trivial Augment, Respectively. Best and Second Best AugReg Settings For Each Model are Highlighted in \textbf{Bold} and \underline{Underline}, Respectively. \textit{Model}$^*$ Represents Results For Image Size 448.}
    
    \label{table_results_augmentations}

\end{table*}

Therefore, we explore the potential behind incorporating generic AugReg techniques developed for coarse image recognition into FGIR tasks. Specifically, we compare four different levels of AugReg: minimal, weak, medium, and strong. The minimal level only employs RC and RF, and is the setting most commonly used in FGIR works \cite{hu_see_2019, rao_counterfactual_2021}. The weak level, which is used in newer works \cite{he_transfg_2022, wang_feature_2021}, also applies LS. The medium level further incorporates SD with a probability of 0.1. Finally, the strong level additionally includes RE and TA. We select TA based on qualitative evaluation that indicates its application of less severe distortions to the image compared to AA. The effects of AugReg choice on classification accuracy for the CUB dataset are shown in \Cref{table_results_augmentations}.

We observe that AugReg is indeed a crucial factor for FGIR and can have a significant impact on accuracy, often rivaling the effect of image size and the choice of model. On the CUB dataset, we observe an average absolute accuracy improvement of 1.59\% and 1.43\% for image sizes of 224x224 and 448x448, respectively, when using strong AugReg compared to minimal AugReg. For reference, the average absolute accuracy improvement from increasing the image size from 224 to 448 on the CUB dataset is 1.97\%.

Since incorporating stronger AugReg incurs minimal increase in training cost while not increasing inference cost at all we suggest practitioners to consider stronger generic AugReg as a cost-efficient method to improve recognition performance.

\subsection{Extended Ablations}

\subsubsection{Importance of Aggregator for Crop Robustness}

To demonstrate the importance of the Aggregator module to our system, we conduct an ablation utilizing crops obtained via random selection. Results are shown in \Cref{table_ablation_random_crop}. As seen in the second row, incorporating random crops without the Aggregator module results in a decrease in performance compared to the baseline, as the random crops deteriorate the quality of input data. Through the usage of our Aggregator module, the method obtains robustness to incorporating random crops, as the self-attention mechanism allows for dynamic reweighting of the contribution of the crops vs the original image. Therefore, when the crops are sub-optimal, the model can effectively discard noise and contributions from the crops.

On the other side, when we guide the crop selection through the proposed GLS metric and the quality of the crop improves, the Aggregator modules incorporates this information to decrease the classification error by 8.29\%. These results, along with the ones presented in \Cref{ssec_ablations} illustrate the importance of both the GLS module for discriminative feature selection and the Aggregator module for high-level feature refinement.

\begin{table}[htb]
    \centering

    \small{
    \begin{tabularx}{\columnwidth}{l Y}
        \toprule
        Baseline (ViT B-16) & $90.11\pm{0.24}$\\
        \midrule
        Random Crops W/O Aggregator & $89.47\pm{0.19}$\\
        \midrule
        Random Crops With Aggregator & $90.26\pm{0.27}$\\
        \midrule
        GLSim With GLS Crops & $\mathbf{90.93\pm{0.24}}$\\
        \bottomrule
        \end{tabularx}
    }

    \caption{Ablation on importance of Aggregator module for robustness to non-optimal crops.}
    \label{table_ablation_random_crop}
\end{table}


\subsubsection{Hyperparameter $O$}

We investigate the impact of the hyperparameter $O$, which governs how many tokens with the highest similarity are employed for crop selection, on the classification performance, in \Cref{table_ablation_o_hparam}. Smaller values lead to more aggressive cropping that may fail to crop the object of interest, while higher values may return a high percentage of background. Despite this, the findings indicate that the proposed approach is relatively robust to variations in this selection criterion, as all improve upon the baseline by at least 5.87\% in the case of $O=4$ and up to 8.29\% in the case of $O=8$.

\begin{table}[htb]
    \centering

    \small{
    \begin{tabularx}{\columnwidth}{Y Y Y Y Y}  
        \toprule
         4 & 8 & 16 & 32\\
        \midrule
        $90.69\pm{0.25}$ & $\mathbf{90.93\pm{0.24}}$ & $90.85\pm{0.17}$ & $90.82\pm{0.30}$\\
        \bottomrule
        \end{tabularx}
    }

    \caption{Ablation on $O$ hyperparameter for cropping on CUB.}
    \label{table_ablation_o_hparam}
\end{table}

\subsubsection{Similarity Metric}

We study the effects of similarity metric choice in \Cref{table_ablation_sim_metric}. We were inspired to employ cosine similarity due to its relation to self-attention, both involving dot products, but these results show that our our method exhibits robustness to the metric selection, as all reduce the ViT baseline relative classification error by at least 8.63\%.

\begin{table}[htb]
    \centering

    \small{
    \begin{tabularx}{\columnwidth}{Y Y Y}  
        \toprule
         Cosine & L1 & L2\\
        \midrule
        $90.93\pm{0.24}$ & $\mathbf{91.03\pm{0.21}}$ & $90.93\pm{0.13}$\\
        \bottomrule
        \end{tabularx}
    }

    \caption{Ablation on similarity metric for selecting crops on CUB.}
    \label{table_ablation_sim_metric}
\end{table}

\subsubsection{Vision Transformer's Patch Size}

We evaluated the proposed system using two distinct patch sizes in \Cref{table_ablation_patch_size}. It is worth remarking that the kernel and stride size utilized in the ViT's patchifier are critical factors that directly influence the granularity of the feature maps, albeit at the expense of increased computational overhead, resulting from the greater effective sequence length.

Results show that the relative classification error compared to the baseline can be reduced by up to 17.23\% for the model with patch size 32. These demonstrate the efficacy of the proposed approach at various feature map granularities.

\begin{table}[htb]
    \centering

    \small{
    \begin{tabularx}{\columnwidth}{l Y Y}
        \toprule
        Patch Size & Baseline (Vanilla ViT) & GLSim\\
        \midrule
        16 & $90.11\pm{0.24}$ & $\mathbf{90.93\pm{0.17}}$\\
        \midrule
        32 & $85.84\pm{0.17}$ & $\mathbf{88.28\pm{0.02}}$\\
        \bottomrule
        \end{tabularx}
    }

    \caption{Ablation on vision transformer patch size on CUB.}
    \label{table_ablation_patch_size}
\end{table}

\section{Additional Samples on GLS For Visualization}
\label{appendix_sec_discussion and future prospects}


\begin{figure*}[htb]
    \centering
    
    \includegraphics[width=0.95\linewidth]{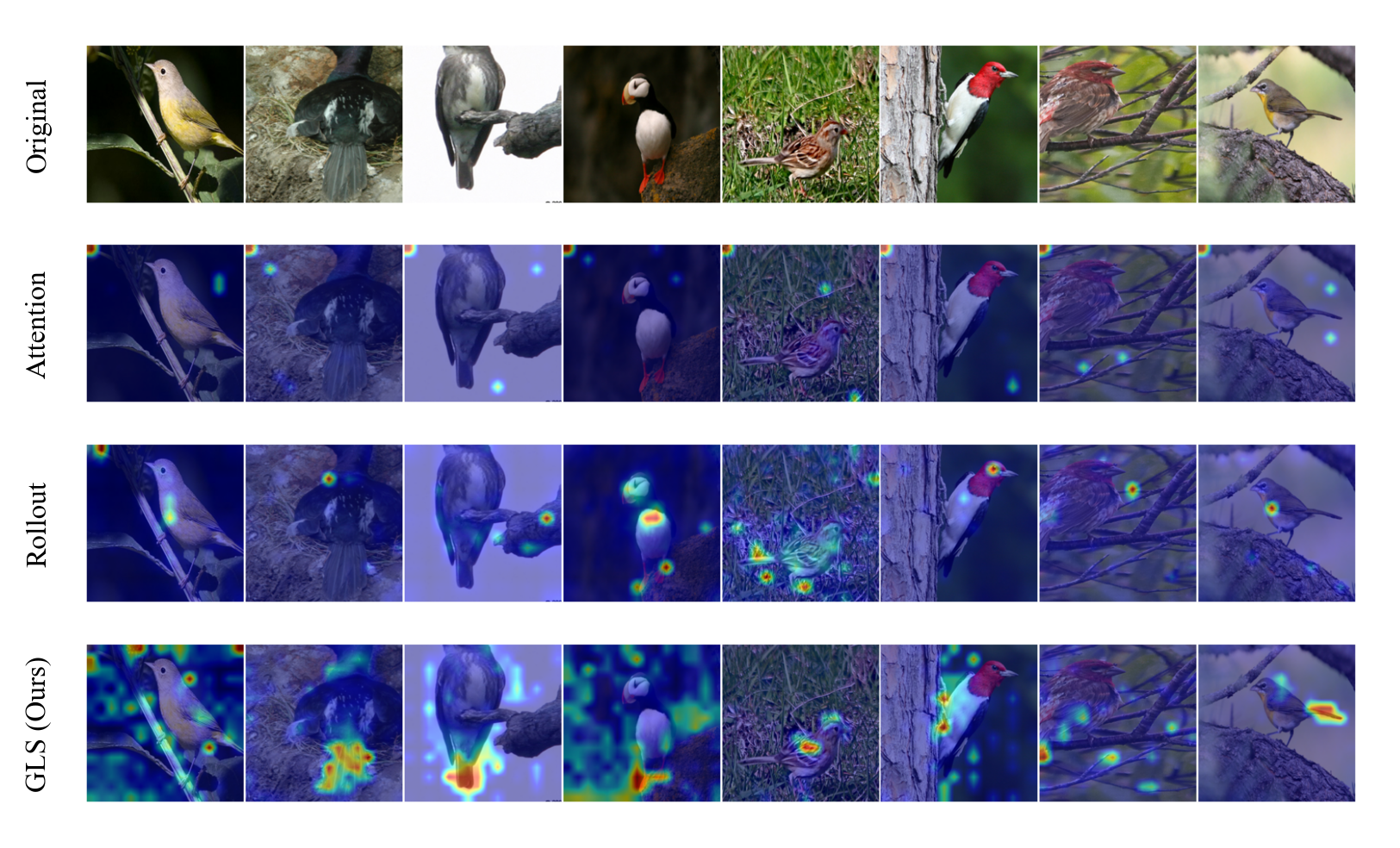}

    \caption{Visualization of attention-based visualization mechanisms and our proposed GLS for DINOv2 B-14 \cite{oquab_dinov2_2023} on CUB.}
    \label{figure_dfsm_dinov2_cub}
\end{figure*}



\begin{figure*}[htbp]
    \centering
    
    \includegraphics[width=0.95\linewidth]{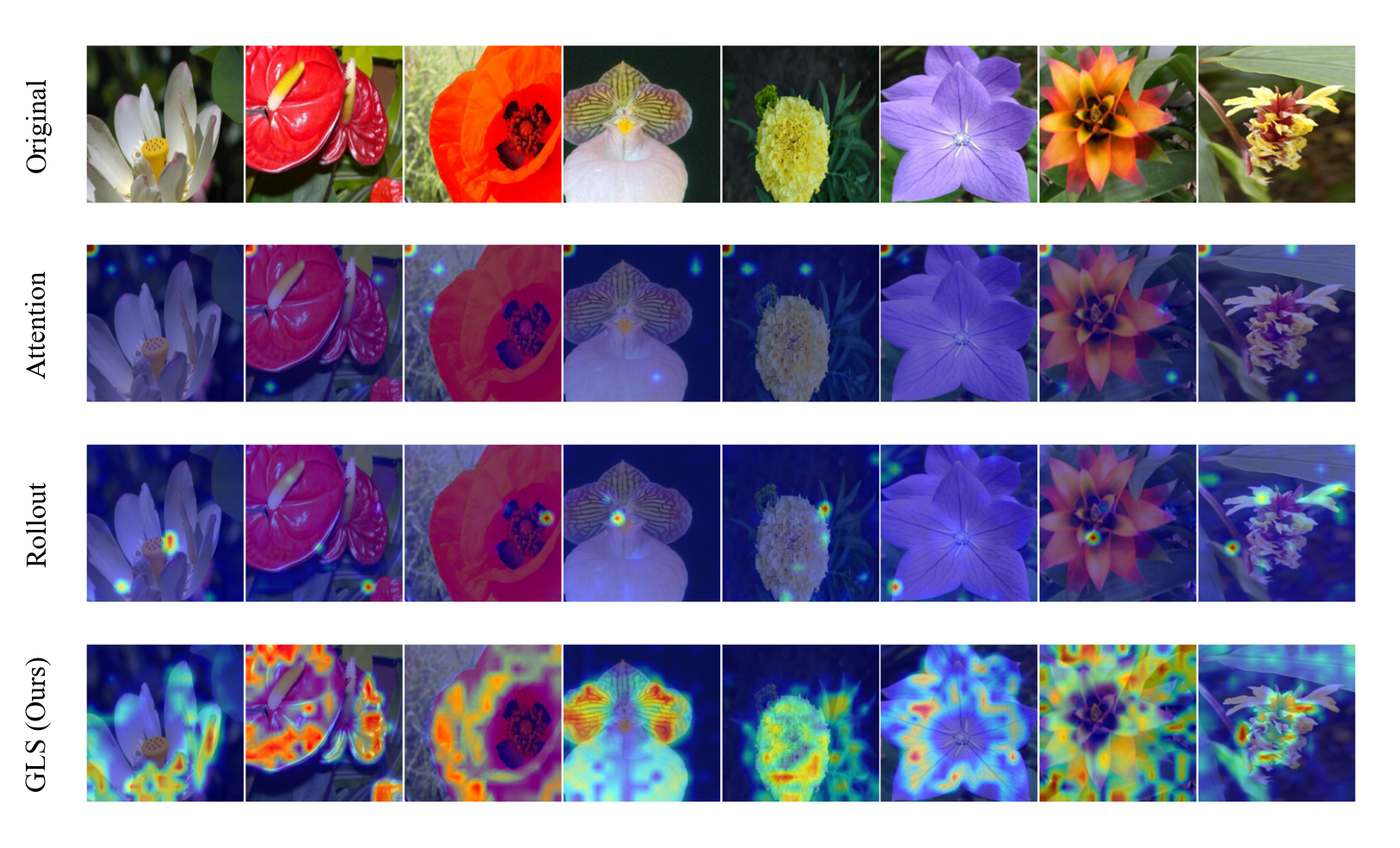}

    \caption{Visualization of attention-based visualization mechanisms and our proposed GLS for DINOv2 B-14 \cite{oquab_dinov2_2023} on Flowers.}
    \label{figure_dfsm_dinov2_flowers}
\end{figure*}



\begin{figure*}[htbp]
    \centering
    
    \includegraphics[width=0.95\linewidth]{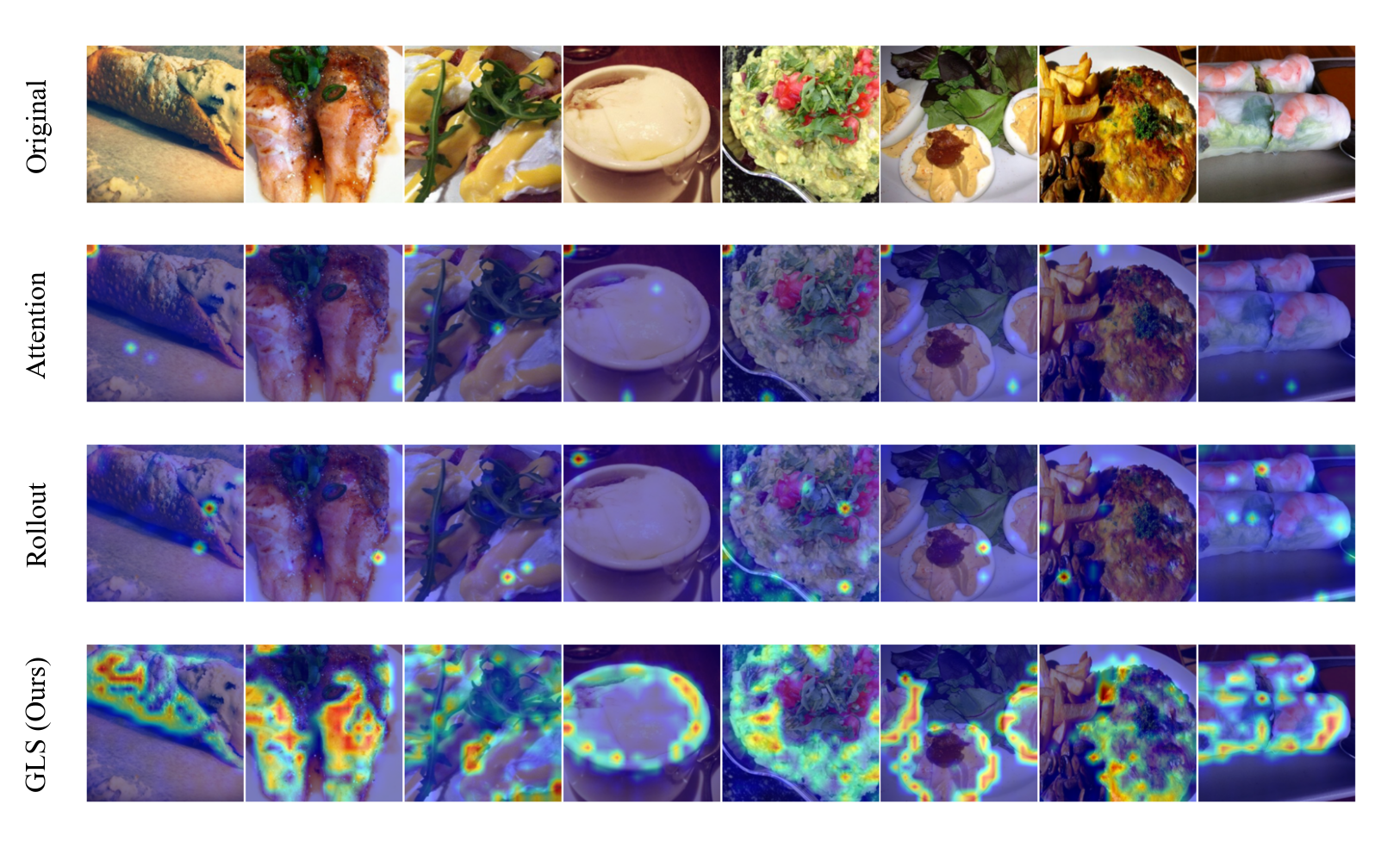}

    \caption{Visualization of attention-based visualization mechanisms and our proposed GLS for DINOv2 B-14 \cite{oquab_dinov2_2023} on Food.}
    \label{figure_dfsm_dinov2_food}
\end{figure*}



\begin{figure*}[htbp]
    \centering
    
    \includegraphics[width=0.95\linewidth]{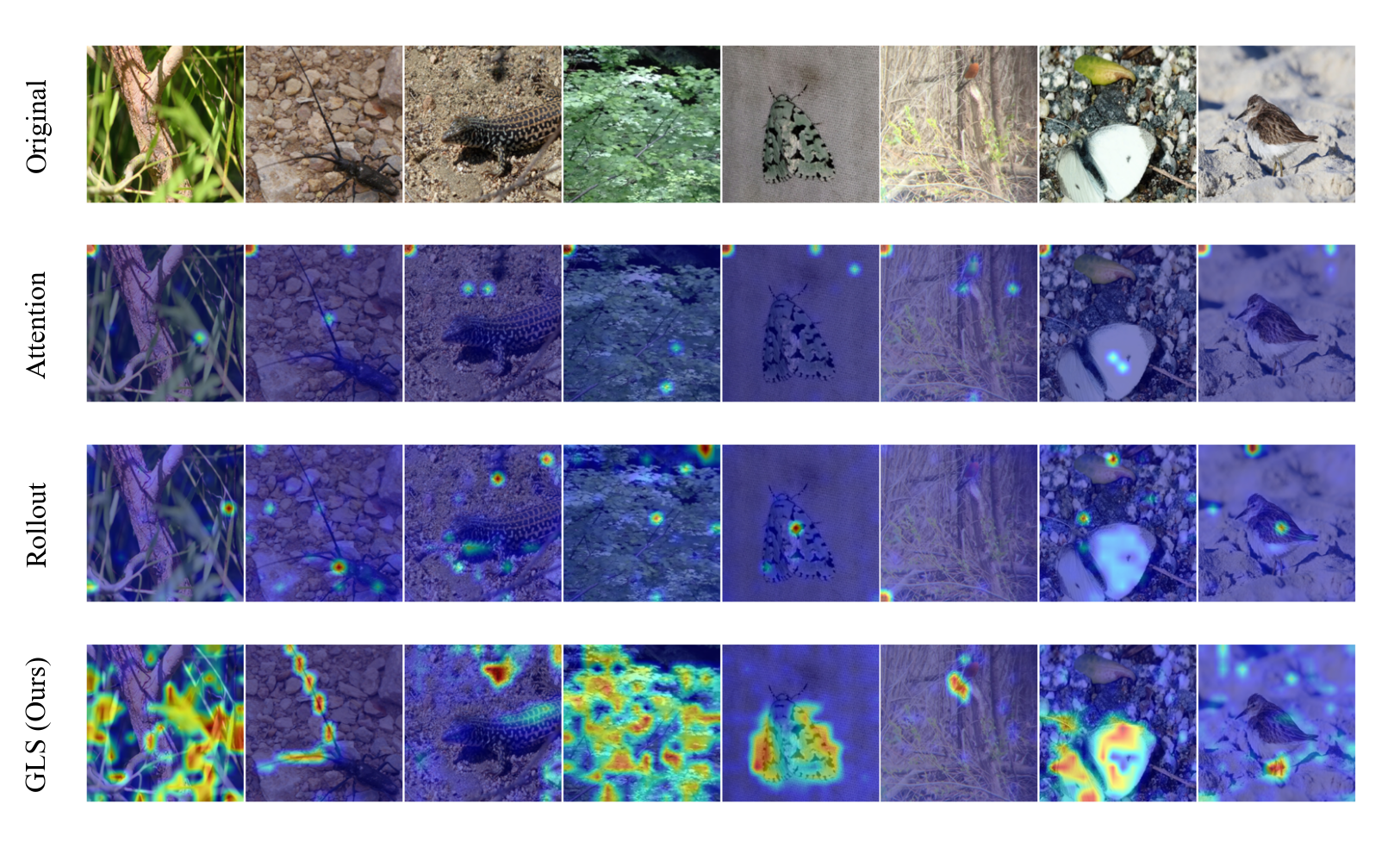}

    \caption{Visualization of attention-based visualization mechanisms and our proposed GLS for DINOv2 B-14 \cite{oquab_dinov2_2023} on iNat17.}
    \label{figure_dfsm_dinov2_inat17}
\end{figure*}



\begin{figure*}[htbp]
    \centering
    
    \includegraphics[width=0.95\linewidth]{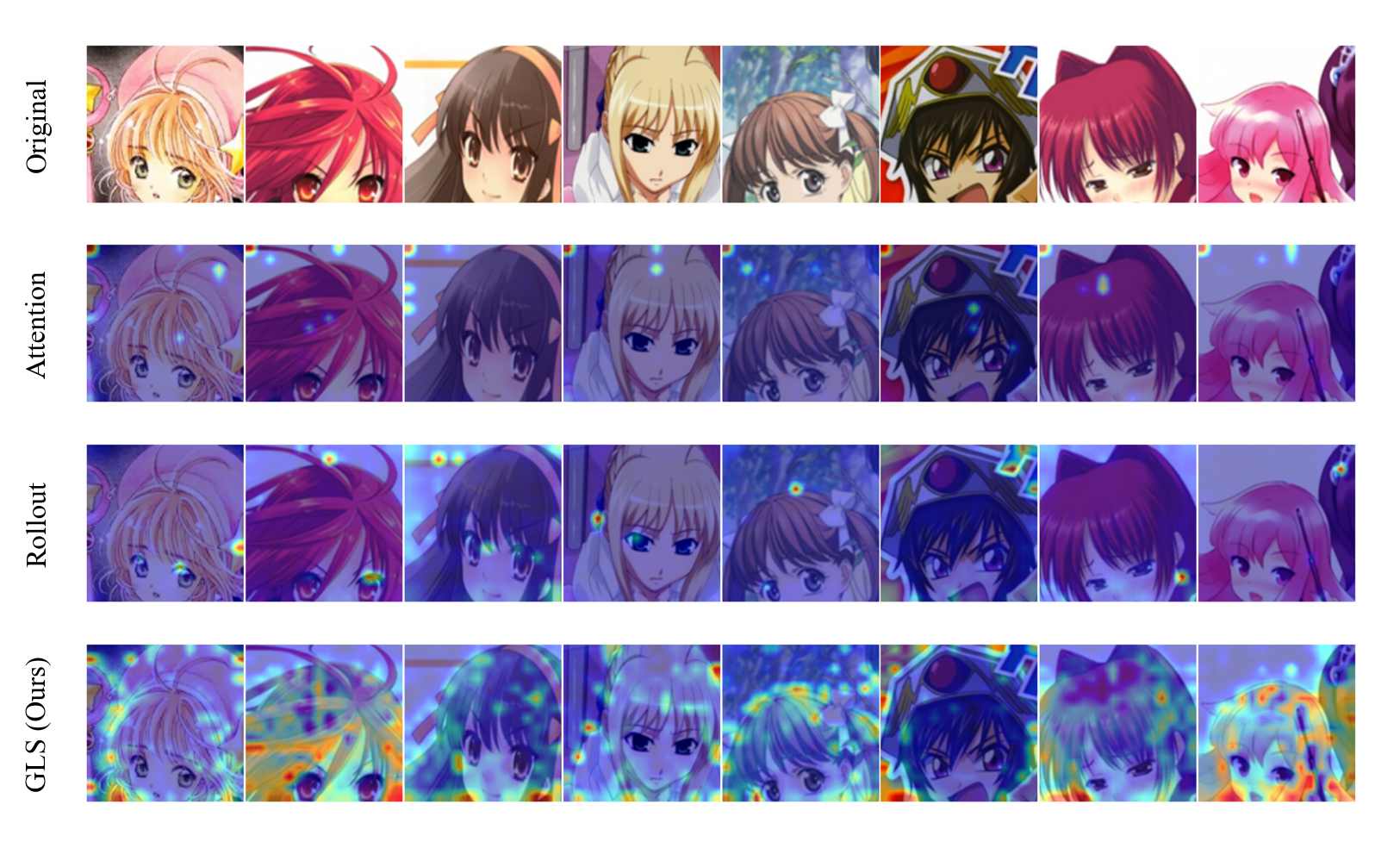}

    \caption{Visualization of attention-based visualization mechanisms and our proposed GLS for DINOv2 B-14 \cite{oquab_dinov2_2023} on Moe.}
    \label{figure_dfsm_dinov2_moe}
\end{figure*}



\begin{figure*}[htbp]
    \centering
    
    \includegraphics[width=0.95\linewidth]{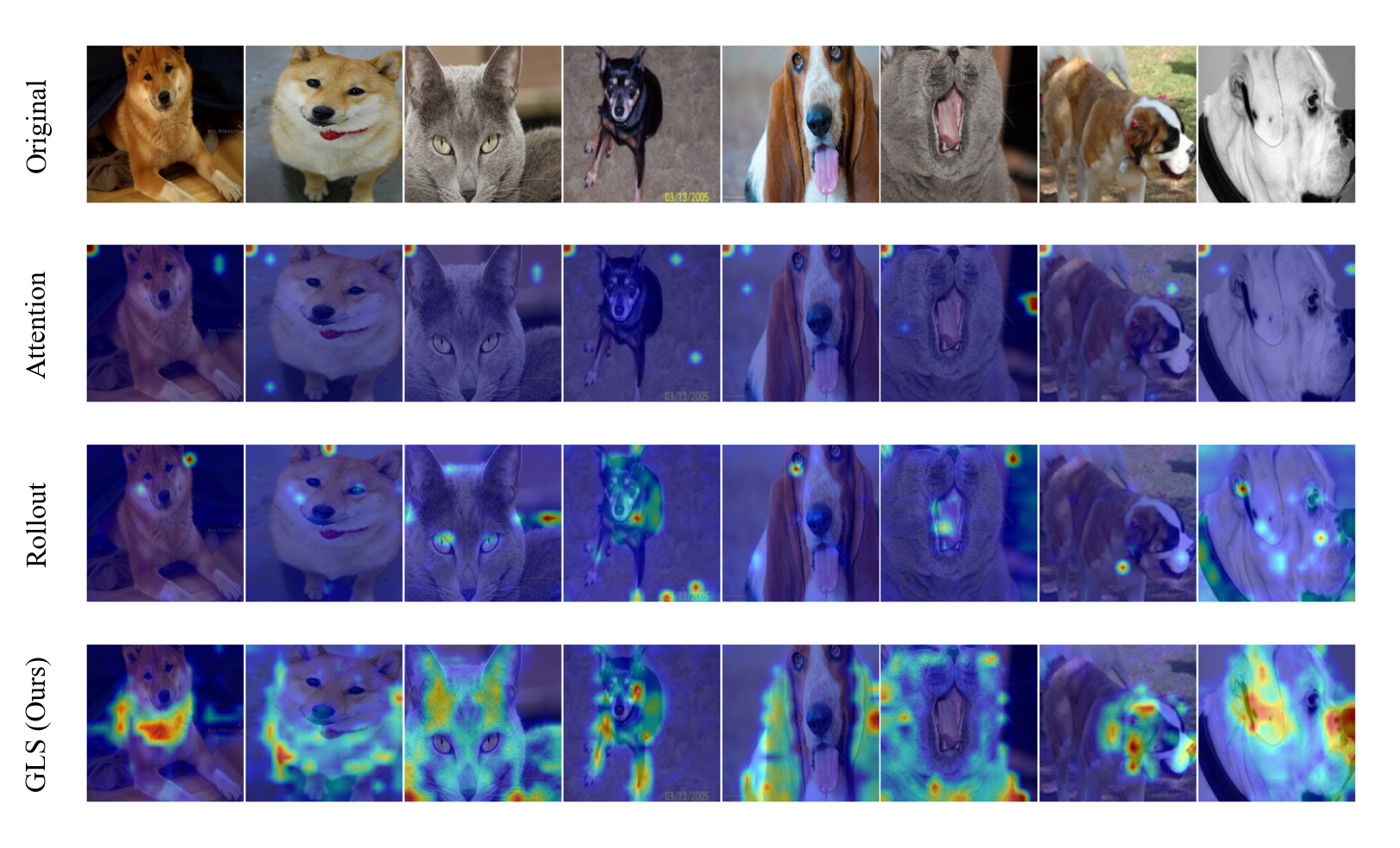}

    \caption{Visualization of attention-based visualization mechanisms and our proposed GLS for DINOv2 B-14 \cite{oquab_dinov2_2023} on Pets.}
    \label{figure_dfsm_dinov2_pets}
\end{figure*}



\begin{figure*}[htbp]
    \centering
    
    \includegraphics[width=0.95\linewidth]{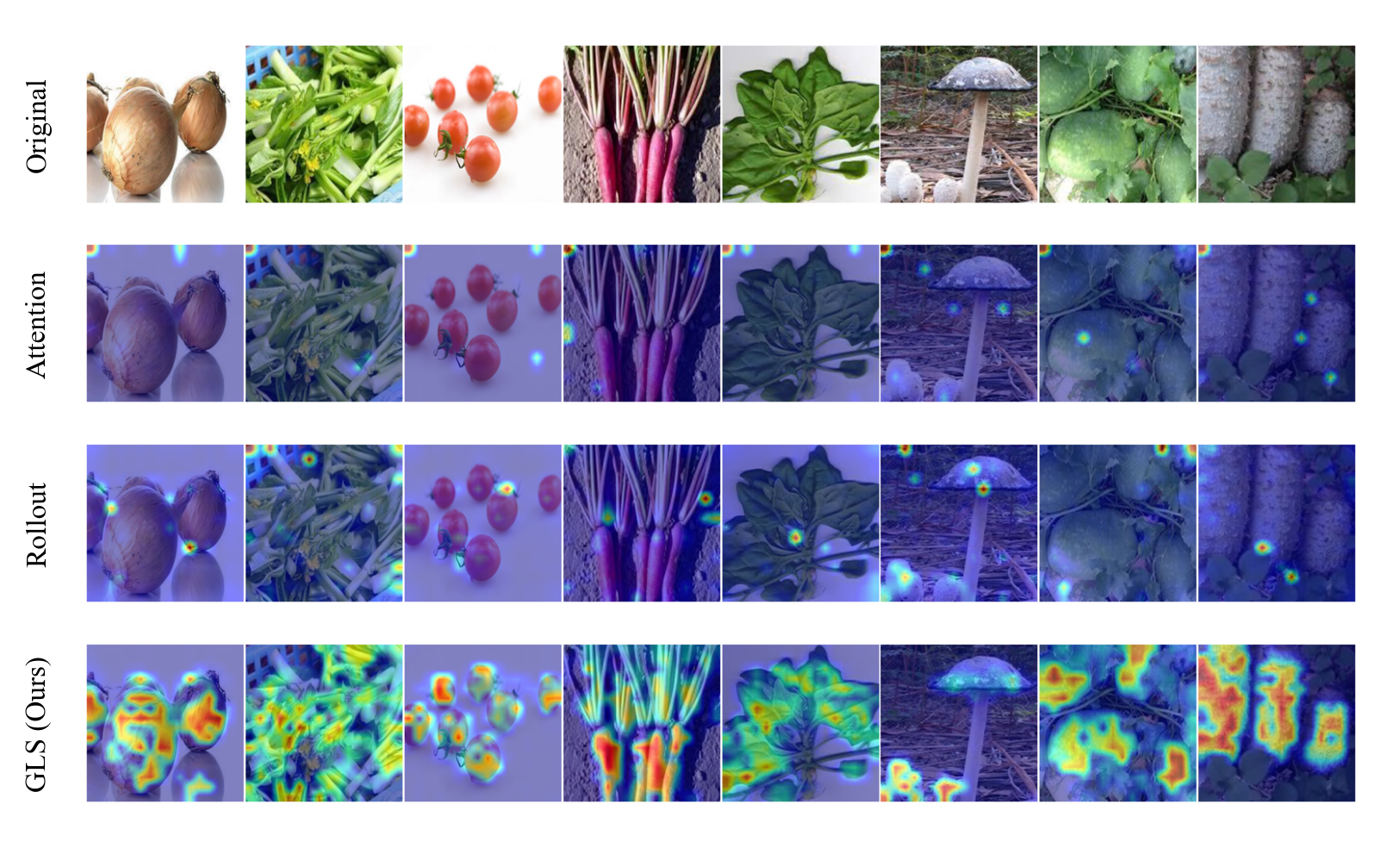}

    \caption{Visualization of attention-based visualization mechanisms and our proposed GLS for DINOv2 B-14 \cite{oquab_dinov2_2023} on VegFru.}
    \label{figure_dfsm_dinov2_vegfru}
\end{figure*}


We include additional samples of the proposed GLS metric as a visualization tool to highlight which local regions of the image have high similarity to the global representation from the CLS token to interpret classification predictions when combined with state-of-the-art pretrained backbones DINOv2 B-14 \cite{oquab_dinov2_2023} across different datasets in \Cref{figure_dfsm_dinov2_cub} to \Cref{figure_dfsm_dinov2_vegfru}.

\end{document}